%% file: main_arxiv.tex
\documentclass[runningheads]{llncs}

\usepackage{eccv}

\usepackage{eccvabbrv}

\input{preamble}

\newcommand{\vn}[1]{\mathbf{#1}}

\usepackage{multirow}
\usepackage{enumitem}
\usepackage{makecell}
\usepackage{amsmath}
\usepackage{amssymb}
\usepackage{graphicx}
\usepackage{subcaption}
\usepackage{gensymb}
\usepackage{booktabs}
\usepackage{algorithm}
\usepackage{algpseudocode}
\usepackage{textcomp}
\usepackage{adjustbox}
\usepackage{wrapfig}

\input{macros}

\usepackage[accsupp]{axessibility}  

\usepackage{orcidlink}

\usepackage{hyperref}

\usepackage{orcidlink}

\begin{document}

\title{Domain Reduction Strategy \\ for Non-Line-of-Sight Imaging} 


\author{
Hyunbo Shim\inst{*}\orcidlink{0009-0001-6408-1992} \and
In Cho\inst{*}\orcidlink{0009-0006-2131-4430}\and
Daekyu Kwon\orcidlink{0009-0006-2154-3348} \and
Seon Joo Kim\orcidlink{0000-0001-8512-216X}
}

\authorrunning{H. Shim et al.}

\institute{Yonsei University}

\maketitle

\def\thefootnote{*}\footnotetext{Equal contribution.}
\def\thefootnote{\arabic{footnote}}

\input{Main/Sections/1_abstract}
\input{Main/Sections/2_intro}
\input{Main/Sections/3_related_work}
\input{Main/Sections/4_method}
\input{Main/Sections/5_experiment}
\input{Main/Sections/6_discussion}
\input{Main/Sections/7_conclusion}


%
%
\bibliographystyle{splncs04}
\bibliography{main}

\clearpage
\section*{Supplementary Material}
\input{Supple/supple}

\end{document}

%% file: preamble.tex
\usepackage[dvipsnames]{xcolor}


%% file: macros.tex
\newcommand{\Tref}[1]{Table~\ref{#1}}

\newcommand{\Eref}[1]{Equation~(\ref{#1})}

\newcommand{\Fref}[1]{Fig.~\ref{#1}}

\newcommand{\Sref}[1]{Section~\ref{#1}}

\newcommand{\Aref}[1]{Algorithm.~\ref{#1}}

\newcommand{\textblock}[1]{\noindent\textbf{#1}}

\def\eg{\emph{e.g.}\ } 
\def\ie{\emph{i.e.}\ }

\def\etal{\emph{et al.}\ }

\newcommand{\parabf}[1]{\paragraph{\textbf{\upshape #1}}}

\def\eg{\emph{e.g}\onedot} 

\def\ie{\emph{i.e}\onedot}

\def\etal{\emph{et al}\onedot}


%% file: Main/Sections/1_abstract.tex
\begin{abstract}
This paper presents a novel optimization-based method for non-line-of-sight (NLOS) imaging that aims to reconstruct hidden scenes under general setups with significantly reduced reconstruction time.
In NLOS imaging, the visible surfaces of the target objects are notably sparse.
To mitigate unnecessary computations arising from empty regions, we design our method to render the transients through partial propagations from a continuously sampled set of points from the hidden space.
Our method is capable of accurately and efficiently modeling the view-dependent reflectance using surface normals, which enables us to obtain surface geometry as well as albedo.
In this pipeline, we propose a novel domain reduction strategy to eliminate superfluous computations in empty regions.
During the optimization process, our domain reduction procedure periodically prunes the empty regions from our sampling domain in a coarse-to-fine manner, leading to substantial improvement in efficiency.
We demonstrate the effectiveness of our method in various NLOS scenarios with sparse scanning patterns.
Experiments conducted on both synthetic and real-world data support the efficacy in general NLOS scenarios, and the improved efficiency of our method compared to the previous optimization-based solutions. Our code is available at \url{https://github.com/hyunbo9/domain-reduction-strategy}.

\keywords{Computational imaging \and Non-line-of-sight imaging}
\end{abstract}

%% file: Main/Sections/2_intro.tex
\section{Introduction}

The ability to see objects placed over the direct line-of-sight has the potential to advance various applications, \eg, autonomous driving, medical care, and rescue operations. Non-line-of-sight (NLOS) imaging aims to reconstruct hidden scenes from measurements of indirect reflections.
These transient measurements are commonly obtained via ``looking around the corner'', where a light source and a time-of-flight sensor illuminate and scan the relay wall (see \Fref{fig:intro} (a)).

\input{Main/Partials/figure_intro}
Recent advancements in NLOS imaging have been achieved by FFT-based inverse solutions, including 3D convolution-based methods 
\cite{o2018lct, young2020dlct, ahn2019convolutional}, and wave-based methods \cite{lindell2019fk, liu2020diffraction}.
Despite promising results, one or more assumptions are made to model the exact solutions, \eg, planar relay walls or certain scanning systems.
These methods also necessitate a dense scanning procedure to reconstruct high-resolution outputs, which is too time-consuming for practical uses.

Considering the practical applications of NLOS imaging, there are numerous scenarios where the exact solutions are not sufficient to meet the demands.
In such scenarios, an ideal relay wall may not be available, and dense scanning of the relay wall becomes impractical.
There is also a need to reconstruct precise surface geometry.
These scenarios give rise to the need for solutions capable of solving NLOS imaging under broader settings.
Several works have attempted to meet this demand with optimization-based NLOS methods \cite{heide2019partial, tsai2019optimization, shen2021netf, pei2021dynamic}.
Albeit versatile, they require substantial computational resources per iteration, with a significant portion dedicated to the unoccupied regions of the hidden scenes.

In this paper, we present a novel optimization-based method that inherits the versatility of previous optimization-based solutions while simultaneously enhancing efficiency. 
In NLOS imaging scenarios, the visible surfaces of the target objects are notably sparse, occupying less than 5\% regions of the entire hidden space.
Since the empty regions do not actually contribute to both transients and the target volumes, we aim to exclude these unnecessary computations.

Pruning the empty regions requires processing varying size inputs.
Therefore, we tailor our method to render the transients by partially propagating the light returning from a set of an arbitrary number of points.
We begin by modeling the transients as a superposition of point-wise light propagation functions.
Given a set of points sampled from the hidden space, light propagation of each point is computed using its albedo and surface normal.
The predicted transients are obtained by a linear combination of the computed propagations, optimized to reconstruct the target volumes through inverse rendering.
Our pipeline is accurately and efficiently implemented through a matrix-free CUDA kernel.
This facilitates our method to accurately model both surface geometry and albedo, without being constrained to specific types of relay walls and scanning systems.

Once we model the transients as a superposition of point-wise propagations, superfluous computations for the empty regions can be easily removed.
Here, we propose a novel domain reduction strategy to eliminate such unnecessary computations, resulting in a notable improvement in the efficiency of the optimization process.
During the optimization process, our domain reduction strategy periodically identifies regions of which albedo becomes lower than a certain threshold and prunes these empty regions from our sampling domain.
Sampling a set of points only from the active regions effectively boosts the efficiency of our method, achieving about $20 \times$ acceleration of the reconstruction time.
This domain reduction procedure is conducted in a coarse-to-fine manner, enabling our method to reconstruct high-resolution output volumes with a single commercial GPU.

We demonstrate the validity of our method in various NLOS scenarios, including a non-planar relay wall, sparse scanning patterns, confocal and non-confocal, and surface geometry reconstruction (\Fref{fig:intro} (b)).
Our experimental results from both synthetic and real-world inputs confirm that our method can efficiently reconstruct hidden objects with fine details across various scenarios. Notably, our method is capable of producing $128 \times 128$ outputs in under a minute.

%% file: Main/Partials/figure_intro.tex
\begin{figure}[t]
\centering
\includegraphics[trim={0.0pt 30 0 0}, width=1\linewidth]{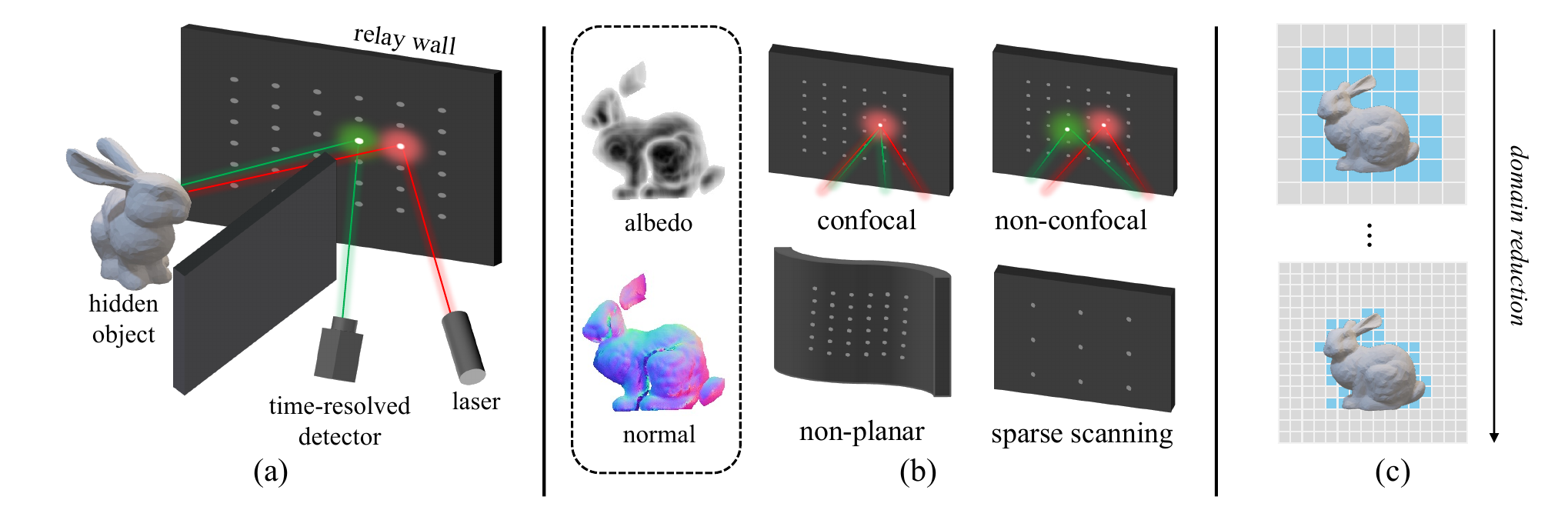}
\caption{
\textbf{(a)} A common NLOS scanning system. A laser and a time-resolved sensor illuminate and scan the relay wall. \textbf{(b)} Our method reconstructs both albedo and surface normal of the hidden objects in general scenarios, including non-confocal, non-planar relay walls and sparse sampling. \textbf{(c)} Our domain reduction gradually prunes empty regions in a coarse-to-fine manner, achieving significant efficiency improvement.
}
\label{fig:intro}
\end{figure}

%% file: Main/Sections/3_related_work.tex
\section{Related Work}

\textblock{NLOS imaging.}
NLOS imaging was first introduced by Kirmani \etal\cite{kirmani2009looking}, and experimentally validated in \cite{velten2012fbp} by using a femtosecond laser and a streak camera. Various methods have been developed for NLOS imaging, which can be broadly divided into two parts: active \cite{o2018lct, lindell2019fk, isogawa2020optical, tsai2017first, isogawa2020efficient, xin2019fermat, choi2023self, zhu2022remapping, wang2023non} and passive methods \cite{seidel2020two, tanaka2020polarized, saunders2019computational, boger2019passive, batarseh2018passive, yedidia2019using, bouman2017turning}.
Passive methods rely on indirect light under various setups, \eg, using videos of a blank wall \cite{sharma2021what}, a large transmissive window \cite{shi2024passive}, or recovering videos projected in a hidden scene \cite{aittala2019mirrors}.
On the other hand, active NLOS methods utilize controllable a light source and a time-of-flight detector, usually producing higher-quality results with the use of more abundant information.
In the active NLOS setup, the pulsed laser emits photons to the visible wall and measures the amount of returning photons to reveal the shape of the hidden object.
The active NLOS imaging methods has been rapidly enhanced with various input sources, such as acoustic waves \cite{lindell2019acoustic}, and diverse types outputs, \eg, albedo \cite{o2018lct, ahn2019convolutional}, normal \cite{young2020dlct, grau2022occlusion}, surface \cite{tsai2019optimization}, and depth \cite{chopite2020deep}.

\textblock{Inverse NLOS methods.}
Significant development has been recently made in NLOS imaging through various inverse solutions.
By assuming isotropic light scatters and a planar relay wall, LCT \cite{o2018lct} demonstrates that the NLOS imaging problem can be interpreted as a 3D deconvolution problem under the confocal setup.
Their main concept is further extended by DLCT \cite{young2020dlct}, which uses a vector deconvolution to obtain both the surface normal and albedo.
Lindell \etal\cite{lindell2019fk} present a wave-based solution named f-k migration.
Phasor field methods \cite{liu2019phasor, liu2020diffraction} formulate the NLOS imaging problem as a diffractive wave propagation and utilize the Rayleigh-Sommerfeld integral.
Despite impressive results, they require several assumptions, \eg planar relay wall, specific (\eg, confocal) scanning system and exhaustive raster scanning for high-resolution outputs. In addition, methods except DLCT \cite{young2020dlct} can only reconstruct the albedo of hidden scenes.
Although such constraints allow faster computations, general applicability and accurate surface modeling are sacrificed in most inverse NLOS methods.

\textblock{NLOS methods for general setups.}
A number of studies have explored NLOS imaging algorithms that can be applied in various environments, \eg diverse acquisition patterns and geometry \cite{ye2021compressed, liu2023sscr, gu2023fast, rapp2020edge, isogawa2020c2nlos}, and imaging over long ranges \cite{wu2021over}.
Several works utilize additional hardware for various purposes, such as SPAD arrays for faster scanning \cite{nam2021low, pei2021dynamic, mu2022rescue}, which are often expensive for real-world applications, and new hardware designs for dynamic relay walls \cite{la2020non}.
Back-projection (BP) solvers \cite{velten2012fbp, arellano2017fast, ahn2019convolutional} are one of the commonly used techniques for general NLOS imaging.
Iterative optimization-based methods \cite{tsai2019optimization, plack2023fast, iseringhausen2020non} could be effective solutions for NLOS imaging in challenging real-world scenarios.
Ahn \etal\cite{ahn2019convolutional} propose the iterative approach that can operate without being bound to a certain scan system and the lateral resolution of the transients.
Heide \etal\cite{heide2019partial} propose a volumetric-based optimization method that can handle partial occlusion and surface normal.
The optimization-based method using a point spread function \cite{pei2021dynamic} was also proposed.
Tsai \etal\cite{tsai2019optimization} propose the surface optimization method that can reconstruct the continuous surfaces of the hidden scenes, of which results are sensitive to the initial state.
NeTF \cite{shen2021netf} proposes neural representations similar to NeRF \cite{mildenhall2020nerf} for NLOS imaging. 
Importantly, most of them suffer from huge computational costs of the optimization, usually $O(N^5)$ or $O(N^6)$ per iteration.
This paper aims to inherit the general applicability of the optimization-based methods and to address expensive computation overheads of the optimization by identifying and pruning unnecessary computations.

%% file: Main/Sections/4_method.tex
\section{Method}

\input{Main/Partials/figure_method}
\subsection{NLOS Measurement Model}
We begin by formulating the transient imaging model for an arbitrary measurement environment.
In the active NLOS methods, a time-resolved sensor measures returning photons after light pulse is emitted onto a relay wall.
Let $\vn{p}$ be a point in the hidden volumes which follows Lambert's cosine law. We assume no interactions between the surfaces, including inter-reflections and self-occlusion. The measured transients can be formulated as
\begin{equation}
   \begin{aligned}
        & \; \tau(t, \vn{l}, \vn{s}) = \int_{\vn{p} \in \Omega} \rho(\vn{p}) \cdot \Phi_{\vn{p}}(\vn{l}, \vec{n}(\vn{p})) \cdot \Upsilon_{\vn{p}}(\vn{l}, \vn{s}) \cdot \delta(d_\vn{l}+d_\vn{s}-tc) \; \vn{dp}, \\
        & \quad \Phi_{\vn{p}}(\vn{l}, \vec{n}(\vn{p})) = 
        \langle {\vn{l}-\vn{p} \over ||\vn{l} - \vn{p}||}, \vec{n}(\vn{p}) \rangle, \; 
        \Upsilon_{\vn{p}}(\vn{l}, \vn{s}) = {1 \over d_\vn{l}^2} \cdot {1 \over d_\vn{s}^2},
    \end{aligned} 
\label{eq:problem}
\end{equation}
where $\vn{l}$ and $\vn{s}$ denote the laser and scan point at the relay wall, and $\rho(\vn{p})$ and $\vec{n}(\vn{p})$ are the albedo and the surface normal of point $\vn{p}$ in the hidden space $\Omega$.
$\Phi_\vn{p}(\vn{l}, \vec{n})$ models the view-dependent cosine term according to Lambert's cosine law.
$\Upsilon_\vn{p}(\vn{l}, \vn{s})$ models the distance fall-off, where $d_\vn{l}$, $d_\vn{s}$ are the distances from the hidden scene to the laser and the scan point respectively.
The Dirac delta $\delta(\cdot)$ relates time to the light travel distance, $c$ is the speed of light and $t$ is the arrival time of photons.
Note that our formation model is defined without assumptions on relay walls and scanning systems.
By additionally assuming a confocal scanning system and a planar relay wall, our formation model becomes equivalent to the formation model of DLCT \cite{young2020dlct}.

Since each point $p$ in $\Omega$ independently contributes to the transients in our formation model, we can represent the transients as a superposition of functions that models the light propagation from each $p$.
To decompose the transients into a set of point-wise propagation functions, we first model the light propagation function for given $\vn{p}$:
\begin{equation}
g_\vn{p}(t, \vn{l}, \vn{s}) = \Phi_\vn{p}(\vn{l}, \vec{n}(\vn{p})) \cdot \Upsilon_\vn{p}(\vn{l}, \vn{s}) \cdot \delta(d_\vn{l} + d_\vn{s} - tc).
\label{eq:propagation}
\end{equation}
The propagation function $g_\vn{p}$ determines the arrival time, distance fall-off and cosine factor according to $\vn{l}$ and $\vn{s}$.
It can be interpreted as a tailored point spread function that depends on $\vn{p}$, $\vn{l}$, $\vn{s}$ and the surface normal $\vec{n}(\vn{p})$ at $\vn{p}$.
With the above equation, the entire transients can be expressed as
\begin{equation}
    \tau = \int_{\vn{p} \in \Omega} \rho(\vn{p})\ \cdot\ g_\vn{p}\ \vn{dp}.
\label{eq:superposition}
\end{equation}
Each $\rho(\vn{p}) \cdot g_\vn{p}$ represents 3D sub-signals containing all measured photons returning from point $\vn{p}$. 
Our goal becomes revealing $\rho$ and $\vec{n}$ to $p$ that correspond to each decomposed function, without posing constraints on $\vn{l}$, $\vn{s}$ and without losing accurate modeling of light propagation.

\subsection{Reconstruction Scheme}
By ignoring $\Phi$ and $\Upsilon$, the above equation can be addressed using an elliptical Radon transform \cite{liu2019analysis}.
It would enable easier approaches to find the inverse, but the accurate modeling would also be sacrificed.
Therefore, we aim to reconstruct $\rho$ and $\vec{n}$ of the hidden volumes through the optimization.
Instead of using a discretized matrix with a fixed size, we leverage a set of an arbitrary number of points as inputs, which are continuously sampled from the hidden space.

\Fref{fig:method} (a) illustrates the overview of our proposed reconstruction scheme.
We first divide the hidden space into a grid and assign 4-dimensional variables, albedo $\rho$ and normal $\vec{n}$ to each vertex of the grid. In each optimization step, input points are sampled from the hidden space. Then the light propagation of each point is computed, which are composed to obtain the transient predictions. 
The 4-dimensional variables are optimized using gradient descent by minimizing the difference between the rendered and the ground truth measurements.
We illustrate details of our optimization scheme in the following paragraphs.

\textblock{Input points sampling.}
Instead of using fixed points as inputs, we randomly sample a set of continuous points, from the hidden space at every step.
We utilize a grid-based random sampling technique to obtain a spatially balanced points set. We sample one point from each cell of the 3D voxel grid, resulting in the set of points $H$.
The albedo and normal at each point are determined by the trilinear interpolation of the surrounding 8 vertices.
The normals are converted to unit vectors after the interpolation.

\textblock{Light propagation and superposition.}
To synthesize the transients with a sampled points set $H$, we discretize the transients in \Eref{eq:superposition} to a finite linear combination of $\{G_\vn{p}\}$, where $G_\vn{p}$ can be expressed as
\begin{equation}
    \begin{gathered}
        G_\vn{p} = \Phi_\vn{p} \cdot \Upsilon_\vn{p} \cdot 1_{A}, \cr 
        A = \{ (t, \vn{l}, \vn{s}) \; | \; d_\vn{l} + d_\vn{s} = tc \}.
    \end{gathered}
\label{eq:discretization}
\end{equation}
To compute $G_\vn{p}$, we first identify the non-zero histograms of each scan point $\vn{s}$ with given $\vn{l}$ and $\vn{s}$, by computing $t$ that satisfies the indicate function $1_A$.
The contribution at the corresponding location is then determined by $\Phi_\vn{p}$ and $\Upsilon_\vn{p}$, which are calculated as in \Fref{fig:method} (b).
Computations of light propagation are done in parallel for all points in $H$, and thus can be accelerated by GPU.
Finally, we compute the predicted transients with a linear summation:
\begin{equation}
T = \sum_{\vn{p} \in H} \rho(\vn{p}) \cdot G_\vn{p} \cdot \Delta V,
\label{eq:synthesize}
\end{equation}
where $\Delta V$ is volume of a grid bin.
We update $\rho$ and $\vec{n}$ using the gradient descent, by minimizing the L2 distance between rendered and ground truth transients.

\textblock{Noise regularization.}
The sensor noise present in NLOS imaging could adversely affect to our optimization process. To mitigate this issue, we utilize L1 regularization to the albedo to encourage sparsity, and jointly optimize additional noise parameters $d(t, \vn{l}, \vn{s})$ to cover the effects of the background noise, including both ambient light and dark counts.
Specifically, these learnable parameters are defined and assigned for each transient histogram $\tau(t, \vn{l}, \vn{s})$.
To constrain the range of these parameters, and thereby prevent them from covering too much intensities of the rendered transients,
we model the noise parameters $d(t, \vn{l}, \vn{s})$ as
\begin{equation}
    d(t, \vn{l}, \vn{s}) = b + \lambda \sigma(z(t, \vn{l}, \vn{s})),
\end{equation}
where $b$ is the base noise level controlling the minimum values of the noise parameters, $\sigma$ denotes the sigmoid activation function, and $z$ is the learnable parameters that are jointly optimized throughout the optimization process.
$\lambda$ is the hyperparameter controlling the maximum values of $d(t, \vn{l}, \vn{s})$.
These simple techniques greatly improves robustness of our method to the noise.

\subsection{Domain Reduction Strategy}
One crucial observation is that most of the hidden volumes to reconstruct is empty in the NLOS imaging scenarios, since photons are only reflected from the visible surfaces of the objects.
These empty regions do not actually affect either the measured transients or the reconstruction volumes.
By taking the advantages of this sparsity nature, most of the computations in synthesizing transients can be eliminated, as our reconstruction pipeline is capable of computing transients with an arbitrary number of points.
The synthesized transients in \Eref{eq:synthesize} can be divided as
\begin{equation}
    \begin{gathered}
        T = \sum_{\vn{p} \in \Omega_1} \rho(\vn{p}) \cdot G_\vn{p} \cdot \Delta V + \sum_{\vn{p} \in \Omega_2} \rho(\vn{p}) \cdot G_\vn{p} \cdot \Delta V, \cr
        \Omega_1 = \{ \; \vn{p} \in \Omega' | \; \rho(\vn{p}) > \epsilon \; \}, \Omega_2 = \Omega' \setminus \Omega_1,
    \end{gathered}
\label{eq:divide}
\end{equation}
where $\Omega' \subset \Omega$ is a set of points that corresponds to the vertices in the grids.
Since the values in the summation are always positive, we can ignore the operations of the second term if $\epsilon$ is a sufficiently small value ($\vert \vert \rho(\vn{p}) \cdot \ G_\vn{p} \vert \vert_{\infty} \rightarrow 0 $ as $ \epsilon \to 0$ $\forall \vn{p} \in \Omega_2$).
Hence, we can exclude $\Omega_2$ from our domain and eliminate unnecessary computations on $\Omega_2$.
This domain reduction procedure is periodically conducted during the optimization, significantly accelerating the efficiency of our method.

\textblock{Soft domain reduction.}
We observed that the albedo $\rho$ often takes on incorrect values during the optimization, which can result in the accidental removal of non-empty regions from the domain.
To this end, we propose a soft domain reduction strategy that gently reduces the domain with a low-pass filter.
Specifically, we apply a Gaussian kernel to the albedo volumes and then we exclude the grid bins with values smaller than a certain threshold from the domain.
This soft domain reduction expands the domain to the surrounding areas and provides additional chances to be optimized for the accidentally removed regions.

\input{Main/Partials/algorithm1.tex}
\textblock{Coarse to fine strategy.}
Predicting transients with a large number of sampled points is computationally exhaustive, while the coarse shapes of the hidden objects can be sufficiently discovered with coarsely sampled input points.
Therefore, we initialize the hidden space with a coarse grid and gradually increase the resolution to progressively reconstruct fine details.
After sufficiently reducing the domain, we expand the grid to a higher resolution. The grid bins within the active regions are subdivided into smaller bins, and variables of the divided bins are assigned values obtained with trilinear interpolation of previous coarse grids.
Reconstructing in a coarse-to-fine fashion, which shares similar philosophy with various studies in 3D computer vision \cite{liu2020nsvf, yu2021plenoctrees, martin2021nerf, shen2021netf}, enables our method to efficiently reconstruct high-resolution outputs with fine details.

\textblock{Overall algorithm.}
Equipped with the above equations and techniques, our method can be described as in \Aref{al:method}.
The optimization process fully exploits the parallel nature of our formulations with GPU implementations.

%% file: Main/Partials/figure_method.tex
\begin{figure*}[t]
\centering
\includegraphics[trim={0 45 0 0pt}, width=1\linewidth]{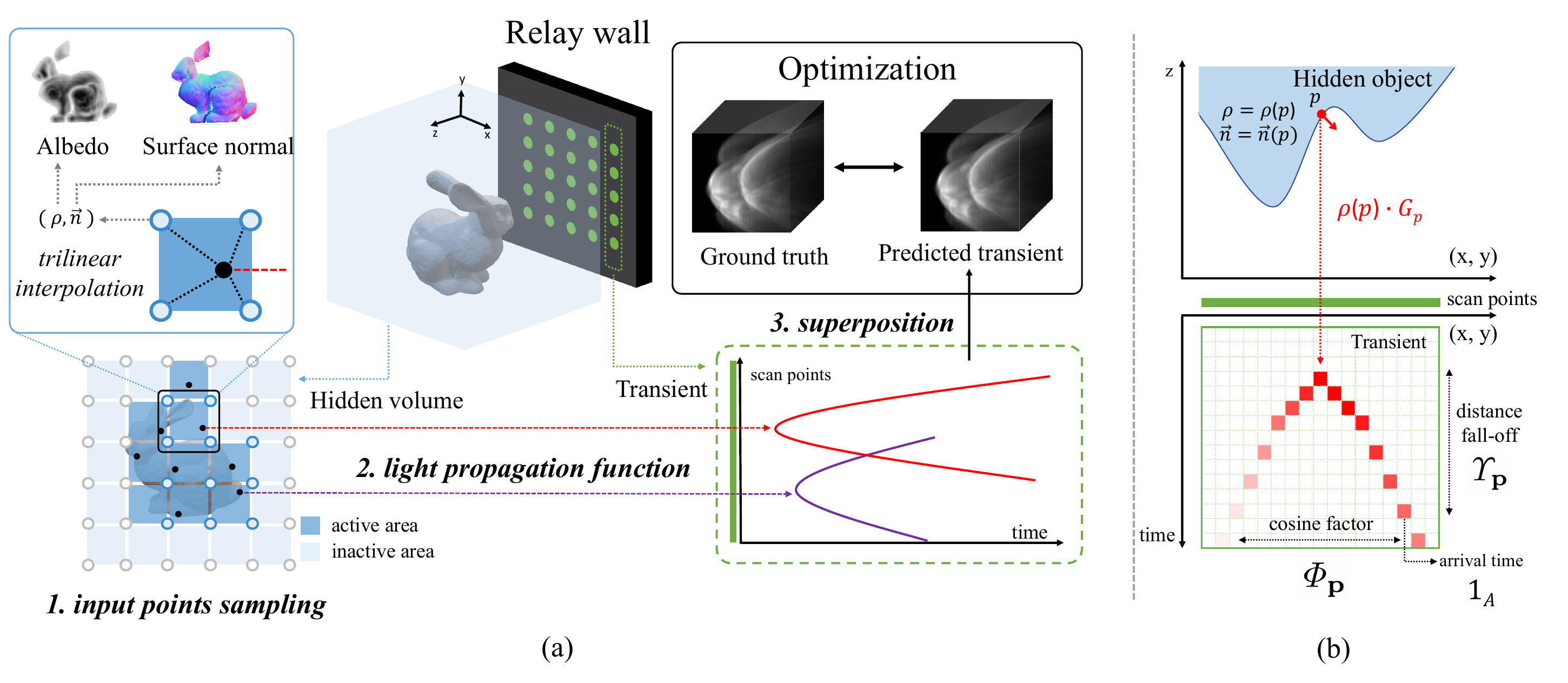}
\caption{
\textbf{(a)} The overview of our reconstruction scheme.
We divide hidden space to a grid shape and assign albedo $\rho$ and surface normal $\vec{n}$ for each vertex.
Input points are randomly sampled from the hidden space. Then the light propagation from each point $G_\vn{p}$ is computed, of which superposition is used as predicted transients. The variables are optimized by minimizing the L2 distance. 
Our domain is gradually reduced by pruning the empty regions during the optimization.
\textbf{(b)} The computation of light propagation function from each point $\vn{p}$. The arrival time at scan point $\vn{s}$ is first identified, and then the fall-off terms $\Phi_\vn{p}$ and $\Upsilon_\vn{p}$ are computed using $\vn{p}$, $\vn{l}$, $\vn{\vn{s}}$ and $\vec{n}(\vn{p})$.
}
\label{fig:method}
\end{figure*}

%% file: Main/Partials/algorithm1.tex
\begin{algorithm}[t!]
\caption{Reconstruction Algorithm.}
\begin{algorithmic}[1]
\footnotesize
\State $\Omega' \leftarrow$ Set of points of grid vertices of hidden scene
\For {$step \leftarrow 1, 2, ..., N $}
    \State $S \leftarrow$ Set of variables $\rho$ and $\vec{n}$ on $\Omega'$
    \State $H \leftarrow$ Set of points, one for each bin on $\Omega'$
    \State $T \leftarrow$ Linear combination of $\{G_p\}_{p\in H}$  \Comment{ Equation (5) }
   \State $T \leftarrow$ $T +$ Synthesized noise  \Comment{ Equation (6) }
    \State Update $\rho, \vec{n} \in S$ by minimizing $\lVert T - \tau_{gt} \rVert_2$

    \If {$step\; \in \; reduction\_step$}
        \State $\rho_{\texttt{low}} \leftarrow$ Low-pass filter$( \rho )$
        \State $\Omega' \leftarrow$ Domain reduction($\Omega'$,   $\rho_{\texttt{low}})$ \Comment{ Equation (7) }
    \EndIf

    \If {$step\; \in \; expansion\_step$}
        \State $\Omega' \leftarrow$ Dividing $\Omega'$ into a fine grid
        \State $S \leftarrow$ Trilinear interpolation($S$)
    \EndIf
\EndFor \\
\Return $\rho, \vec{n}$
\end{algorithmic}
\label{al:method}
\end{algorithm}

%% file: Main/Sections/5_experiment.tex
\section{Experiments}
We demonstrate the effectiveness of our method through the extensive experiments.
We compare the results on transients having $32 \times 32$ scannings, with both confocal and non-confocal scanning systems, and with a non-planar relay wall.

\textblock{Baselines.}
We compare our method with several state-of-the-art baselines: FK \cite{lindell2019fk}, DLCT \cite{young2020dlct}, Phasor field \cite{liu2019phasor} with a BP solver (without FFT), Gram \cite{ahn2019convolutional}, and NeTF \cite{shen2021netf}.
We apply maximum intensity projection through the $z$ axis for all methods.
For the results of DLCT, we follow the authors and report the $z$-directional albedo.
We provide results of more baselines in Supplement.

\textblock{Dataset.}
We validate the results on the ZNLOS \cite{galindo19znlos} dataset and the Stanford real-world measurements \cite{lindell2019fk}.
ZNLOS consists of a number of synthetic transients, simulated with a $1\ \textrm{m} \times 1\ \textrm{m}$ relay wall and hidden objects placed approximately $0.5\ \textrm{m}$ from the relay wall. 
The temporal resolution of bins corresponds to the light travel time of $0.3\ \textrm{cm}$.
The Stanford real-world measurements \cite{lindell2019fk} are obtained with a $2\ \textrm{m} \times 2\ \textrm{m}$ relay wall ($1 \textrm{m} \times 1\ \textrm{m}$ for a non-planar case) and a bin resolution $32\ \textrm{ps}$. The raw $512 \times 512$ measurements are first spatially downsampled by 2, which results in $256 \times 256$ transients.
We adopt Bunny and Serapis instances from ZNLOS dataset, and Statue, Dragon, NT instances from the Stanford dataset. 
The NT instance is used for the non-planar evaluation.

\textblock{Implementation detail.}
Our method is implemented using PyTorch, and the light propagation function is implemented using the PyTorch CUDA extension.
We optimize the variables using the Adam optimizer \cite{kingma2014adam} with learning rate 1 and total 1000 steps.
We set the reduction threshold as 5\% (3\% for non-confocal) of the maximum albedo value.
The domain reduction step is performed at every 50 steps.
We refer to Supplement for more implementation details.

\input{Main/Partials/figure_experiment_confocal}
\textblock{Sparse scanning.}
It is worth mentioning that most previous studies \cite{o2018lct,lindell2019fk,liu2019phasor,young2020dlct} have reported results on high-resolution transients obtained by dense scanning.
Such exhaustive procedures require much longer scanning time, which is often difficult in several real-world scenarios.
To demonstrate the applicability of our method in such scenarios, we compare the results on $32 \times 32$ sparsely sampled transients.
These sparsely sampled measurements are obtained by taking the subset of the measurements with appropriate spatial strides.

\subsection{Confocal NLOS Imaging Result}
We first evaluate the performance of our method on sparsely sampled confocal measurements.
As shown in \Fref{fig:experiment_confocal}, results of DLCT contain pale artifacts around the hidden objects, making some objects indistinguishable from these artifacts.
FK \cite{lindell2019fk} often fail to reveal fine details of the objects, \eg the ear of Bunny and the head of Dragon.
Phasor \cite{liu2019phasor} with a BP solver yields some artifacts and some details are missing.
Gram \cite{ahn2019convolutional} delivers blurry results or fails to capture several details.
NeTF \cite{shen2021netf} tends to reconstruct the objects with blurry and distorted albedo.
On the other hand, our method successfully reconstructs clean and sharp results with fine details on both synthetic and real-world transients.

\textblock{Normal maps.}
To evaluate the capability of accurately reconstructing surface geometry, we deliver the reconstructed normal maps in \Fref{fig:experiment_normal}.
DLCT is unable to produce accurate surface geometry on $32 \times 32$ transients, indicating the necessity of raster-scanning for the inverse methods.
On the other hand, our method delivers normal maps in high-quality, reconstructing details of the objects, and achieves significantly reduced scanning time compared to DLCT.

\input{Main/Partials/table_quantitative}
\input{Main/Partials/figure_experiment_normal}

\textblock{Quantitative results.}
We also quantitatively evaluate depth maps and normal maps on $32 \times 32$ transients of Bunny.
We upsample the results of all methods to the resolution $256 \times 256$, matching that of the ground truth, with trilinear interpolation.
We also apply 10\% threshold to acquire depth and normal maps.
We measure the mean absolute error (MAE) and root mean squared error (RMSE) for both depth and surface normals.
As reported in \Tref{table:quantitative}, our method outperforms all other baselines in both depths and surface geometry, demonstrating that our method reconstructs accurate geometry as well as visual appearance.

\input{Main/Partials/figure_experiment_various_setting}
\subsection{Various NLOS Imaging Result}
\textblock{Non-planar relay wall result.}
We conduct the experiments on the real-world transients captured with a non-planar wall \cite{lindell2019fk}, to show that our method can be applied to arbitrary types of the sampling geometry.
Following previous works \cite{o2018lct, young2020dlct}, we slightly adjust our light propagation model in this experiment to properly deal with the retro-reflective targets.
We compare the results with FBP \cite{velten2012fbp} and FK \cite{lindell2019fk}.
The wave extrapolation technique is applied to report the results of FK as in their paper.
\Fref{fig:experiment_various_setting} (a) demonstrates the results of non-planar relay wall setup.
FBP is effective at reconstructing the silhouette, but its results tend to be noisy and show some distortions.
FK \cite{lindell2019fk} recovers the blurry shapes of the objects with the incorrect albedo values at some regions such as the shape of N.
Our method reconstructs the clean shapes of the objects with more correct albedo values, showing the extensibility of our method beyond planar relay walls.

\textblock{Non-confocal NLOS imaging result.}
We evaluate the performance of our method with a non-confocal scanning system.
We present our results on the non-confocal $32 \times 32$ transients of Bunny.
Methods designed to address the non-confocal measurements, Phasor \cite{liu2019phasor} and FBP \cite{velten2012fbp}  are chosen and compared as baselines.
\Fref{fig:experiment_various_setting} (b) reports the qualitative results on the non-confocal Bunny.
FBP reconstructs only some parts of Bunny, in which streak artifacts can be observed.
Phasor field fails to recover some details and yields the coarse and blurry output.
Unlike these baselines, our method delivers much cleaner results with more detailed shapes, exhibiting general applicability of our method.

\textblock{Results with various sampling resolutions.}
We report the results on the transients by gradually increasing the sampling resolutions by 2, from $16\times16$ to $64 \times 64$.
We compare our results with DLCT \cite{young2020dlct}, which shares most similar image formation model with our method and FFT-based method.
As in \Fref{fig:experiment_various_setting} (c), our approach delivers plausible results with $16 \times 16$ samplings and already achieves high-quality outputs with $32 \times 32$ samplings.
On the other hand, DLCT fails to produce comparable outputs with our method until the sampling resolution reaches $64 \times 64$.
Interestingly, no meaningful enhancement is observed by increasing the sampling resolution higher than $32 \times 32$, indicating that sparsely sampled $32 \times 32$ scan points are sufficient for our method to achieve high-resolution results, without requiring exhaustive raster-scanning.

\subsection{Analysis and Ablation Study}
In this section, we provide deeper insights and analysis of our proposed concepts. All analysis and ablations are conducted on the confocal $32 \times 32$ transients of Bunny, except for the ablation the noise regularization techniques.

\input{Main/Partials/figure_ablation}
\input{Main/Partials/table_ablation_reduction}
\textblock{Noise regularization.}
We conduct the ablation on our noise regularization techniques to assess the robustness of our method in the presence of sensor noise.
\Fref{fig:ablation} (a) delivers the results on the real-world measurements of Statue and Dragon.
Results of the model without any regularization are severely affected by the presence of noise (No reg.).
Imposing L1 regularization alleviates the effects of the noise, but we can still observe some noisy artifacts (L1 only).
With the jointly optimized noise parameters, our method achieves clean results with much reduced artifacts, showing the robustness of our method to the noise.

\textblock{Domain reduction and runtime analysis.}
We conduct the ablation on our domain reduction strategy to discover the efficiency improvement of our method.
We compare three models, which are without domain reduction, domain reduction without coarse-to-fine strategy, and the proposed method.
We report total runtime and the ratio of remaining active regions at 100, 500, 1000 steps, with the final reconstructed albedo.
As shown in \Fref{fig:ablation} (b) and \Tref{table:ablation_reduction}, our model with coarse-to-fine domain reduction achieves substantial efficiency improvement, without sacrificing the quality of the outputs.
Compared with the model without reduction strategy, reconstruction time of our method is significantly reduced.
The model without coarse-to-fine strategy suffers from heavy computations at the early stage, where the empty regions are not sufficiently pruned yet 100 steps.
Thanks to the coarse-to-fine domain reduction strategy, our method efficiently eliminates most of the unnecessary computations with a low-resolution grid, achieving more than $20 \times$ efficiency improvement as a result. 

\textblock{Accurate light modeling.}
We discover the effects of accurate light modeling through the ablation on the cosine factor and the distance fall-off term.
We compare the results of three variants of our method: optimized without both cosine factor and distance fall-off, model assuming isotropic scatter (no cosine factor), and model with both terms.
As reported in \Fref{fig:ablation} (c), The results of both experiments with less accurate modeling are degraded. The model without both terms shows the worst result, only recovering the silhouette of the bunny.
The model only without cosine term reconstructs almost all details, but still fails to recover some details (circled parts).
These results support that the accurate modeling of light propagation leads to the improvement.

%% file: Main/Partials/figure_experiment_confocal.tex
\begin{figure}[t]
\centering
\includegraphics[trim={0 15 0 0pt}, width=1\linewidth]{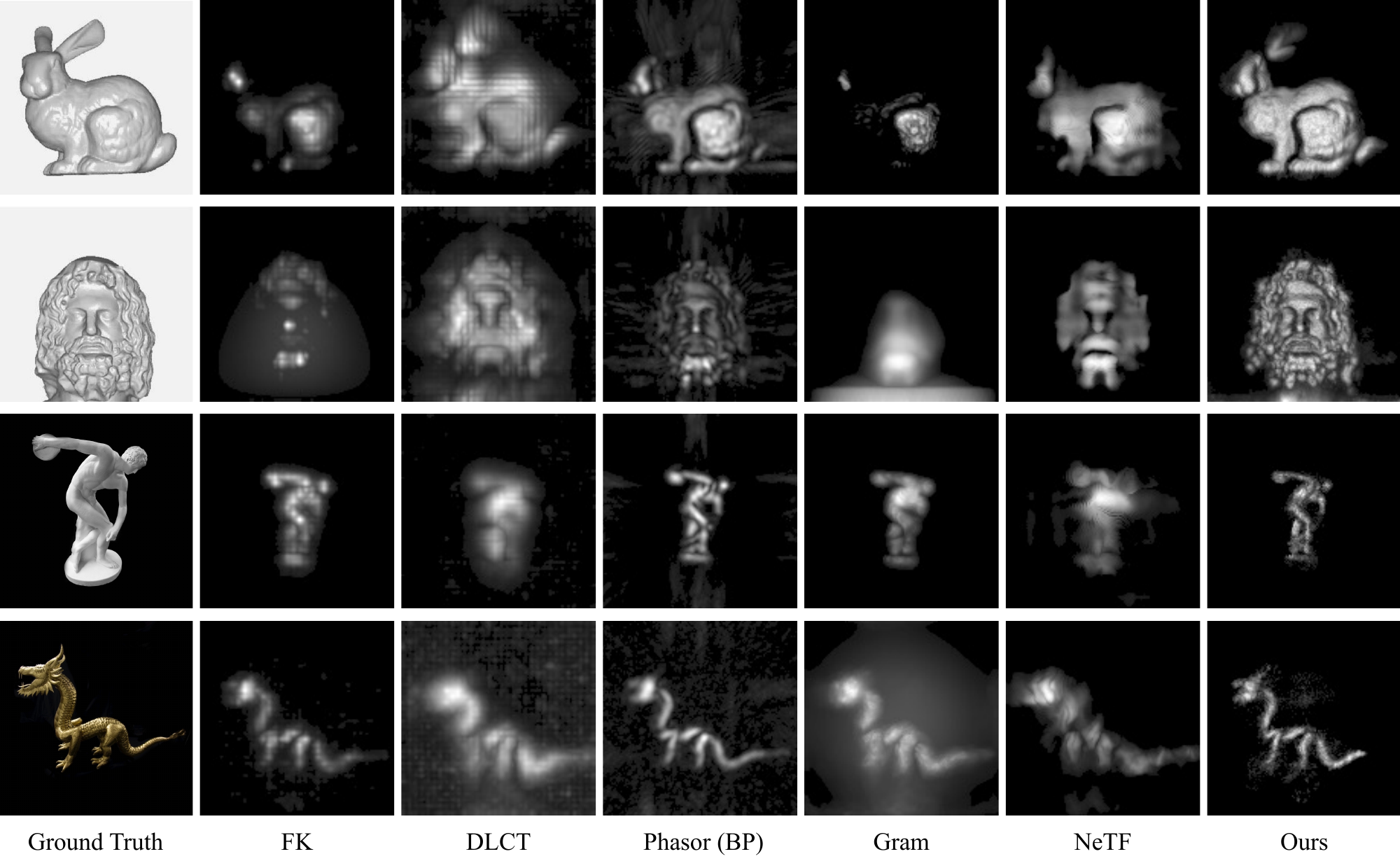}
\caption{
Reconstructed albedo on the transients with a confocal $32 \times 32$ scan system and a planar relay wall. Our method achieves the highest reconstruction quality.
}
\label{fig:experiment_confocal}
\end{figure}

%% file: Main/Partials/table_quantitative.tex
\begin{table}[t!]
\setlength{\tabcolsep}{7pt}
\centering
\small
\caption{
Quantitative results of Depth map and Normal map on the ZNLOS Bunny under confocal setting with threshold values of 5\% and 10\% of maximum albedo.
}

\resizebox{\columnwidth}{!}{

\begin{tabular}{c|cc|cc|cc|cc}
\toprule
\multirow{2}{*}{Method} & \multicolumn{2}{c|}{Depth (5\%)} & \multicolumn{2}{c|}{Normal (5\%)}  & \multicolumn{2}{c|}{Depth (10\%)} & \multicolumn{2}{c}{Normal (10\%)} \\
 & MAE & RMSE & MAE & RMSE  & MAE & RMSE & MAE & RMSE \\
\hline
LCT \cite{o2018lct} & 0.1977 & 0.2875 & - & - & 0.1048 & 0.2069 & - & -  \\
FK \cite{lindell2019fk}& 0.0719 & 0.1871 & - & - & 0.1023 & 0.2173 & - & -  \\
Phasor (BP) \cite{liu2019phasor}& 0.1348 & 0.2049 & - & - & 0.1283 & 0.2068 & - & -  \\
Phasor (FFT) \cite{liu2020diffraction}& 0.1300 & 0.2808 & - & - & 0.0818 & 0.2095 & - & -  \\
Gram \cite{ahn2019convolutional}& 0.0913 & 0.1628 & - & - & 0.0751 & 0.1827 & - & -  \\
NeTF \cite{shen2021netf}& 0.0679 & 0.1748 & - & - & 0.0681 & 0.1754 & - & -  \\
DLCT \cite{young2020dlct}& 0.3189 & 0.4220 & 0.3796 & 0.4856 & 0.1449 & 0.2729 & 0.2021 & 0.3438 \\
Ours & $\bold{0.0477}$ & $\bold{0.1523}$ & $\bold{0.1147}$ & $\bold{0.2394}$ & $\bold{0.0493}$ & $\bold{0.1552}$ & $\bold{0.1115}$ & $\bold{0.2341}$ \\
\bottomrule
\end{tabular}

}

\label{table:quantitative}
\end{table}

%% file: Main/Partials/figure_experiment_normal.tex
\begin{figure}[t]
\centering
\includegraphics[trim={0 10 0 0pt}, width=1\linewidth]{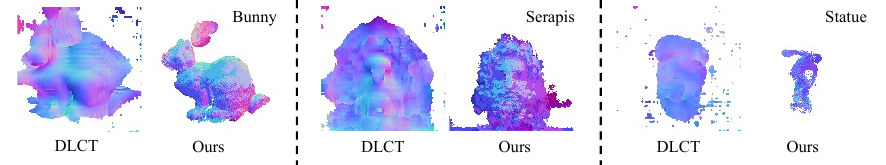}
\caption{
Comparison of the reconstructed normal map with DLCT \cite{young2020dlct} on the transients with a confocal $32 \times 32$ scan system and a planar relay wall. Our method accurately reconstructs surface geometry, whereas DLCT only produces coarse structures.
}
\label{fig:experiment_normal}
\end{figure}

%% file: Main/Partials/figure_experiment_various_setting.tex
\begin{figure}[t]
\centering
\includegraphics[trim={0 10 0 0pt}, width=1\linewidth]{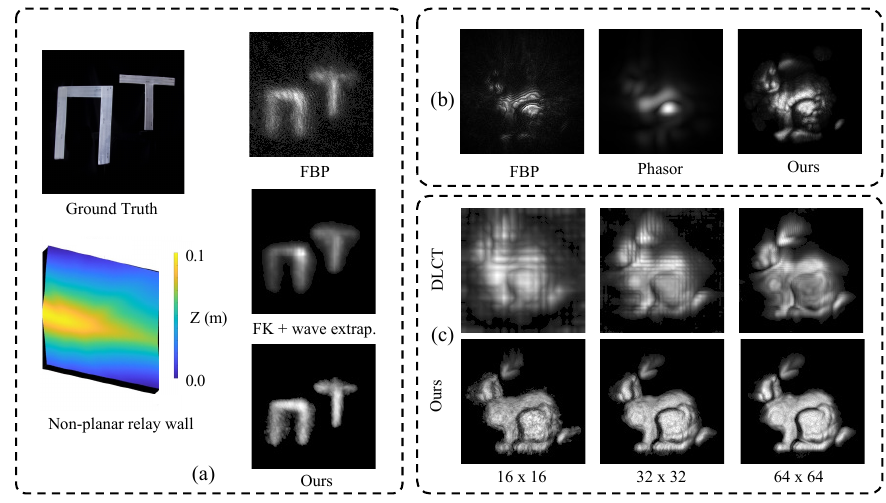}
\caption{
Results on various setting. \textbf{(a)} Results on the real-world $32 \times 32$ transients of the NT instance \cite{lindell2019fk}, measured with a non-planar relay wall. \textbf{(b)} Results on non-confocal ZNLOS Bunny. \textbf{(c)} Results on Bunny with various sampling resolutions.
Our method consistently delivers high-quality results in various scanning scenarios.
}
\label{fig:experiment_various_setting}
\end{figure}

%% file: Main/Partials/figure_ablation.tex
\begin{figure}[t]
\centering
\includegraphics[trim={0 5 0 0pt}, width=1\linewidth]{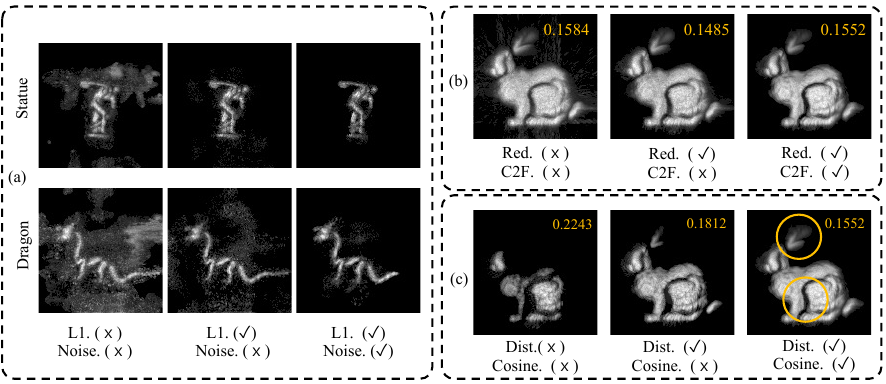}
\caption{
Qualitative and qualitative ablation results.
\textbf{(a)} Ablation results on the noise regularization techniques.
`L1' denotes the L1 regularization and `Noise.' denotes optimized noise parameters.
\textbf{(b)} Ablation results on the reduction. `Red.' denotes the domain reduction and `C2F' denotes the coarse-to-fine strategy. Both `Red.' and `C2F' do not sacrifice the reconstruction quality.
\textbf{(c)} Ablation results on the accurate light modeling. `Dist.' denotes the distance fall-off term and `Cosine.' denotes the cosine factor. The upper-right values denote RMSE of the depth maps.
}
\label{fig:ablation}
\end{figure}

%% file: Main/Partials/table_ablation_reduction.tex
\begin{table}[t]
\begin{adjustbox}{valign=t}
\begin{minipage}{0.5\linewidth}

\setlength{\tabcolsep}{3pt}
\centering
\small
\caption{Ablation results on the domain reduction. `Red.' denotes the domain reduction and `C2F' denotes the coarse-to-fine strategy. We report elapsed time (seconds) of the optimization and the ratio of remaining domain in the parentheses at each step.}
\resizebox{\columnwidth}{!}{
\begin{tabular}{cc|cccc|c}
\toprule
Red. & C2F & 100 iter & 500 iter & 1000 iter \\
\midrule
 &  & 109 s & 539 s & 1087 s \\
\checkmark & & 67 s (11\%) & 103 s (4\%) & 134 s (4\%) \\
\checkmark & \checkmark & 4 s (15\%) & 27 s (3\%) & 54 s (3\%) \\
\bottomrule
\label{table:ablation_reduction}
\end{tabular}
}
\end{minipage}
\end{adjustbox}
\hfill
\begin{adjustbox}{valign=t}
\begin{minipage}{0.47\linewidth}
\setlength{\tabcolsep}{4pt}
\centering
\caption{
Comparisons of GPU memory usage (MB) with FFT-based methods, where $N$ is the hidden volume resolution. Our domain reduction (DR) significantly reduces the consumption especially when the target resolution is high.
We use GPU versions \cite{chen2020lfe} of LCT and FK.
}
\label{table:comparison_memory}
\resizebox{0.95\columnwidth}{!} {
\begin{tabular}{ccccc}
\toprule
\multirow{2}{*}{Resolution} & \multirow{2}{*}{LCT} & \multirow{2}{*}{FK} & Ours & Ours \\
& & & w/o DR & w/ DR \\
\midrule
$N = 32$ & 1473 & 1445 & 1187 & 1211 \\
$N = 128$ & 7101 & 6813 & 4065 & 1659 \\
\bottomrule
\end{tabular}
}
\end{minipage}
\end{adjustbox}
\end{table}

%% file: Main/Sections/6_discussion.tex
\section{Discussion}
\input{Main/Partials/table_general_applicability}
\textblock{Comparison of general applicability.}
We illustrate the general applicability of the NLOS imaging algorithms in \Tref{table:general_applicability}.
As can be seen, none of the methods exhibit general applicability as our method except NeTF \cite{shen2021netf}.
The achievable resolution of all FFT-based inverse methods are bounded to the sampling resolution, which inevitably requires time-consuming scanning procedures.
Although NeTF could operate without being limited to all types of constraints, the training time of NeTF typically exceeds 2 days, limiting the applicability of this method.
Our method, on the other hand, is capable of reconstruct hidden scenes without being limited to scanning resolution, a planar relay wall, and can accurately reconstruct surface normals as well as albedo, typically in a minute.

\textblock{Comparison with FFT-based methods.}
The FFT-based methods exhibit $O(N^3logN)$ complexity and faster reconstruction time than our method, where $N$ is the spatial and temporal resolution.
However, exhaustive raster-scanning procedures required for these methods serve as a major bottleneck, which takes about 46 minutes to achieve $128 \times 128$ outputs on the Stanford dataset.
While FFT-based solutions have faster runtime, our optimization-based pipeline can reconstruct high-quality outputs only using $32 \times 32$ samplings, requiring scanning time less than 3 minutes.
Considering both scanning and reconstruction time, our method can serve as an efficient solution for real-world applications.
We provide more results and discussion on sparse samplings in Supplement.

\textblock{Reconstruction time and memory usage.}
The time complexity of our method $O(c M^2N^3)$ per iteration, where $M$ is the sampling resolution and $c$ is the active ratio.
Since the ratio of active regions is mostly less than 4\%, our domain reduction reduces overall computations about $20 \times$, achieving reconstruction time of about a minute.
Moreover, as shown in \Tref{table:comparison_memory}, our domain reduction effectively reduces memory consumption, particularly when the target resolution is high. This enables our method to efficiently reconstruct hidden scenes using a single commercial GPU, typically  requiring less than 2GB memory.

\textblock{Limitation and future works.}
While our method presented the robustness to noise, we expect that such noise can be more effectively addressed.
Reconstructing the objects with self-occlusion is also a worth exploring subject.
We also expect interesting future research for extending the domain reduction to more complex scenes \eg, scenes with multiple objects, and analyze its behavior.

%% file: Main/Partials/table_general_applicability.tex
\begin{table}[t]
\setlength{\tabcolsep}{7pt}
\centering
\caption{Comparison of the general applicability of various NLOS imaging algorithms.}
\footnotesize
\resizebox{\columnwidth}{!}{

\begin{tabular}{cccccc}
\toprule
Method & Scan system & Sparse scanning & Non-planar & Surface recon. \\
\midrule 
LCT \cite{o2018lct} & Confocal &  & &  \\
FK \cite{lindell2019fk} & Confocal & & $\triangle$ &  \\
DLCT \cite{young2020dlct} & Confocal & & & \checkmark \\
Phasor \cite{liu2019phasor} & - & & &  \\
FBP \cite{velten2012fbp} & - & \checkmark & \checkmark \\
Gram \cite{ahn2019convolutional} & - & \checkmark & &  \\
NeTF \cite{shen2021netf} & - & \checkmark  & \checkmark &  \\
Ours & - & \checkmark & \checkmark & \checkmark \\
\bottomrule
\end{tabular}

}

\label{table:general_applicability}
\end{table}


%% file: Main/Sections/7_conclusion.tex
\section{Conclusion}
This paper presented the new optimization-based method that can address various NLOS imaging scenarios.
Thanks to the domain reduction strategy, conducted in a coarse-to-fine manner, our method can efficiently reconstruct hidden scenes in high-quality, less than a minute, significantly reducing acquisition time.
We expect our method to facilitate broader range of real-world applications.

\section*{Acknowledgements} 
This work was supported by the Samsung Research Funding Center (SRFC-IT2001-04), Artificial Intelligence Innovation Hub under Grant RS-2021-II212068, and Institute of Information \& communications
Technology Planning \& Evaluation (IITP) grant funded by the Korea government(MSIT) (Artificial Intelligence Graduate School Program, Yonsei University, under Grant 2020-0-01361).

%% file: Supple/supple.tex
\appendix

In this supplementary material, we first provide additional discussion and comparison with FFT-based methods in \Sref{sec:supp_fft}. Then, we provide additional details of the proposed method in \Sref{sec:supp_method_detail}, additional comparisons with more baseline methods in \Sref{sec:supp_comp}, reconstruction time of our method in \Sref{sec:supp_recon_time}, and additional evaluations and analysis in \Sref{sec:supp_evaluation}.

\section{Comparison with FFT-based Methods}
\label{sec:supp_fft}
\subsection{FFT-based Methods and Optimization}
We would like to demonstrate that the goal of our work is not to achieve the fastest reconstruction speed.
Instead, this work aims to identify efficiency bottlenecks inherent in previous optimization-based methods, particularly stemming from computations related to empty regions.
Consequently, our domain reduction achieves substantial efficiency improvement (see Table 2 of the main paper), while inheriting the general applicability of the optimization framework.

While FFT-based methods benefit from computational efficiency based on the convolutional theorem or Stolt's method, several assumptions have to be made to achieve this, \eg dense scanning points, planar relay walls, or ignoring surface normals.
Some approximations could relax certain assumptions, but at the cost of sacrificing performance.
In contrast, optimization-based methods are more generally applicable to various scenarios, with the potential for continuous modeling, joint optimization of noise parameters, arbitrary BRDF \cite{shen2021netf}, and sparse samplings \cite{ye2021compressed, liu2023sscr}.
Our domain reduction effectively addresses computational burdens of optimization frameworks, paving the way to unleash their full potential.

\input{Supple/Partials/experiment_resize}
\input{Supple/Partials/experiment_various_resolution}
\subsection{Analysis on Sparse Sampling}
Reconstructing high-resolution volumes from undersampled sparse measurements is an ill-posed problem.
Previous FFT-based methods require measurements to have the same resolution with the target volumes.
One may apply interpolation techniques to upsample the measurements to the desired resolution, as commented by R3.
However, this inevitably introduces approximation errors between upsampled and actual measurements.
Therefore, solutions of all methods exhibit approximation errors, which become larger as scanning patterns become sparser.
Optimization-based methods are capable of finding solutions with minimal errors through iterative minimization procedures.
This is further demonstrated in the following paragraphs.

\parabf{Results on $32 \times 32$ with bicubic interpolation.}
For more precise comparisons, we deliver the results of FFT-based methods, using the additional bicubic interpolation technique.
These results are obtained by first applying the bicubic interpolation to upsample the transients to the $128 \times 128$ spatial resolution, and then applying the FFT-based methods to the upsampled transients.
As shown in \Fref{fig:supp_comp_bicubic}, the additional bicubic interpolation improves the results of 3D convolution-based methods \cite{o2018lct, young2020dlct}, while the results of FK and Phasor field are less affected by the interpolation technique.
Nevertheless, LCT and DLCT still produces blurry outputs, incorrect albedo values (see results of Bunny), and inaccurate fine details of the objects (\eg, the ear of Bunny, the face of Serapis, the head of Dragon).
Contrary to these methods, our method reconstructs hidden volumes directly from $32 \times 32$ transients, reconstructing high-quality volumes with fine details, while being robust to the effects of noise.

\parabf{Results on $16 \times 16$ sampling.}
We also present the results of FFT-based methods on the $16 \times 16$ sparsely sampled measurements, which are upsampled to the lateral resolution of the target volumes by filling unsampled pixels with zero.
As shown in \Fref{fig:various_resolution_in_rebuttal} (a), approximation errors caused by the upsampling become evidently larger as scanning patterns become sparser, leading to unfavorable results of the FFT-based methods.
On the other hand, our optimization-based framework successfully reconstructs clean shapes of the objects with many details in these challenging scenarios.

\parabf{Results on $256 \times 256$ measurements.}
Finally, we deliver the results on $256 \times 256$ measurements of ZNLOS \cite{galindo19znlos} Bunny in \Fref{fig:various_resolution_in_rebuttal} (b). As can be seen, all methods produce compelling results with sufficient scan points and sufficiently long scanning time.

\section{Method Details}
\label{sec:supp_method_detail}

\subsection{Reconstruction Objective}
Combining the noise parameters and the L1 regularization, the objective of our reconstruction pipeline can be described as
\begin{equation}
    \mathcal{L}(\rho, \vec{n}) = || ( T + d ) - \tau_{gt} ||_2 + \alpha || \rho ||_1,
\end{equation}
where $T$ is the rendered transients, $\tau_{gt}$ is the ground truth measurements, and $d$ is the noise parameter defined for each histogram. We set $\alpha$ to 0.8 for real-world scenes and 0.001 for synthetic measurements.

\subsection{Additional Implementation Detail}
In this section, we provide detailed explanations of our implementations for the future reproducibility. 
For albedo variables, we apply the ELU \cite{clevert2015fast} activation function to suppress the negative albedo values.
We set the orthogonal direction from the hidden objects to the relay wall as $(0, 0, -1)$.
With $\vec{n}=(n_x, n_y, n_z)$ representing the surface normal, points of the hidden objects which have positive $n_z$ values (backfaces of the objects) are not visible from the measurements.
To ensure negative $n_z$ values, we apply the tanh to the variables and add $(0, 0, -1)$ to obtain the surface normal vector. This process results in a range $(-1, 1)$ for $n_x$ and $n_y$, and a range $(-2, 0)$ for $n_z$.
Finally, we normalize the surface normal to make it as a unit vector.

Throughout all experiments, our method takes transients with $32 \times 32$ scanning points and reconstructs hidden volumes with a $128 \times 128 \times 333$ resolution, where the last is the resolution along $z$-axis.
The standard deviation of the Gaussian kernel used in the soft domain reduction is set to 3.
We empirically observe that slightly reducing the threshold of the domain reduction under the non-confocal setups yield better reconstruction quality. Therefore, we set the threshold to 5\% for the confocal measurements and 3\% for the non-confocal measurements.
To report the results of fast Fourier transform (FFT)-based methods, we upsample spatial resolution of their outputs with bicubic interpolation to make $128 \times 128$ resolution.
To measure the quantitative results of ZNLOS Bunny, we upsample the spatial resolution of results to $256 \times 256$, matching that of the ground truth.
By analyzing the last 10\% transient histograms along $t$-axis of the real-world measurements, we empirically set $b$ to $0.05$ and $\lambda$ to $0.06$ for the noise regularization.
These values are slightly reduced for reconstructing retro-reflective targets ($b = 0.004$, $\lambda = 0.0012$), which usually have higher maximum intensity values.
To report the results of NeTF \cite{shen2021netf}, we use the original source code provided by the authors, and train this model for 192 epochs with 2 stage training as in the original work.
The training of NeTF takes more than 2 days in our environment.
The optimization process of our method for revealing the albedo and surface normal of a single scene takes 1k iterations, which require about a minute using a single commercial RTX 3090 GPU. 

\input{Supple/Partials/table_runtime_all}
\section{Reconstruction Time}
\label{sec:supp_recon_time}
We deliver the reconstruction time and the ratio of active regions for reconstructing all scenes in \Tref{table:supp_runtime_all}.
Although the reconstruction time could vary according to characteristics of the scenes, \ie remaining domain at each step, our method demonstrates its efficiency across all instances, typically taking about a minute to reconstructing $128 \times 128$ output volumes.
Considering both reconstruction time and scanning time required for the high-resolution outputs, our method can serve as an efficient and effective solution for reconstructing high-resolution volumes with $32 \times 32$ scanning points. 

\section{Additional Comparison}
\label{sec:supp_comp}
To clearly demonstrate the effectiveness of our method, we provide comparisons with additional baseline methods.
These include back-projection (BP) based methods that do not utilize Fourier transform and thus do not suffer from the lateral resolution issues, LCT \cite{o2018lct}, and the fast differentiable renderer \cite{plack2023fast}.

\subsection{Confocal Imaging Result}
We report the results of LCT, Phasor with FFT, FBP with a Laplacian filter (Lap.) and a Laplacian of Gaussian (LoG) filter.
We also deliver the results of Phasor with the wavelength $\lambda = 4\Delta_p$, while the results of Phasor in the main paper are with the wavelength $\lambda = 2\Delta_p$.

\input{Supple/Partials/experiment_fbp_pf}
\input{Supple/Partials/experiment_normal_map}

As reported in \Fref{fig:supp_comp_bp}, our method clearly outperforms all other baseline methods, producing clearer results and successfully recovering fine details. The results of LCT are of low resolution, making it difficult to discern the details of the object.
The results of Phasor with FFT fails to reconstruct specific parts, such as the bunny's ear.
The results of FBP contain streak artifacts and noise, which often make the hidden objects difficult to be identified. 

\subsection{Surface Normal}
We compare the reconstruction results of surface normals with the fast differentiable renderer \cite{plack2023fast}. The differentiable renderer, proposed by Plack \etal, reconstructs colors and surface geometry of the hidden objects through the differentiable rendering pipeline.
As demonstrated in \Fref{fig:normal_map_and_non_planar} (a), our method achieves compelling results in surface normal reconstruction, whereas other baselines, DLCT \cite{young2020dlct} and Plack \etal \cite{plack2023fast}, only reconstruct coarse structures of the surfaces with artifacts, or lack several parts and details of the objects.

\subsection{Non-Planar Relay Wall Result}
We additionally provide comparisons of non-planar relay wall results, with Phasor field with a BP solver. We report the results of Phasor field with two different wavelengths, namely $\lambda = 2\Delta_p$ and $\lambda = 4\Delta_p$.
As shown in \Fref{fig:normal_map_and_non_planar} (b), our method delivers cleanest shapes of the ``NT" instance, while other methods produce noisy results where the shapes of the instance are difficult to identify.

\section{Additional Evaluation}
\input{Supple/Partials/experiment_each_epoch}
\input{Supple/Partials/experiment_depth_map_error}
\label{sec:supp_evaluation}
We provide more evaluation results to clearly demonstrate the effectiveness of our method.
First, we report the results at several steps during the optimization to illustrate that most details of the hidden objects can be reconstructed in the early stages of our optimization process.
Second, we present error map visualizations of the reconstructed depth maps on ZNLOS \cite{galindo19znlos} Bunny.
Then we provide additional analysis and ablation study for a deeper understanding of our method.

\subsection{Results at Different Steps}
In \Fref{fig:supp_each_epoch}, we report the results at several steps (100, 300, 500, 700, 900 steps) during the optimization process.
As can be seen, our method already recovers almost all shapes of the objects at 500th iteration, and finer details are gradually revealed as the iteration progresses. While our method can reconstruct the satisfying outputs at the early stage, we continue the optimization process for high-fidelity results.

\subsection{Depth Map Visualization}
We provide visualizations of error maps of reconstructed depth maps in \Fref{fig:supp_depth_map}. While most of the baseline methods suffer from artifacts or missing details of the objects, our method delivers the accurately reconstructed depth maps while recovering most of the parts of the objects. 

\input{Supple/Partials/experiment_normal_reconstruction}
\subsection{Surface Reconstruction}
We provide surface reconstruction results of our method and DLCT \cite{young2020dlct} in \Fref{fig:surface_reconstruction}. Following \cite{young2020dlct}, we obtain the reconstructed surfaces using the Poisson reconstruction method \cite{kazhdan2006poisson}.
As evident, our method successfully reconstructs detailed surfaces using only $32 \times 32$ scanning points, whereas DLCT yields only coarse shapes of the objects, making it difficult to identify many details.

\input{Supple/Partials/albation_continuous_sampling}
\input{Supple/Partials/ablation_exposure}
\subsection{Additional Analysis}

\parabf{Ablation on continuous points sampling.}
We deliver the ablation results on the continuous points sampling in \Fref{fig:ablation_continous_sampling}. Here, we compare our method, using the continuous grid-based random sampling, with the model using a fixed point sampling, which samples the center of each voxel. As can be seen, the fixed point sampling often produces inaccurate results at some part of the objects (see the circled part in \Fref{fig:ablation_continous_sampling}), while the continuous sampling exhibits more accurate results in both synthetic and real-world datasets.

\parabf{Results with various exposure time.}
To further validate the robustness of our method to noise, we present the results on the confocal real-world measurements with various exposure time. We use the $32 \times 32$ measurements with $28.1\ \textrm{s}$, $56.4\ \textrm{s}$, $168.8\ \textrm{s}$ total exposure time, which correspond to 30, 60, 180 minute total exposure time of the original measurements.
As shown in \Fref{fig:ablation_exposure}, our method exhibits more clean and sharp outputs compare to DLCT \cite{young2020dlct}, showing high-quality results even with $28.1\ \textrm{s}$ total exposure time.

\clearpage

%% file: Supple/Partials/experiment_resize.tex
\begin{figure*}[t]
\centering
\includegraphics[trim={0 0 0 0}, width=1\linewidth]{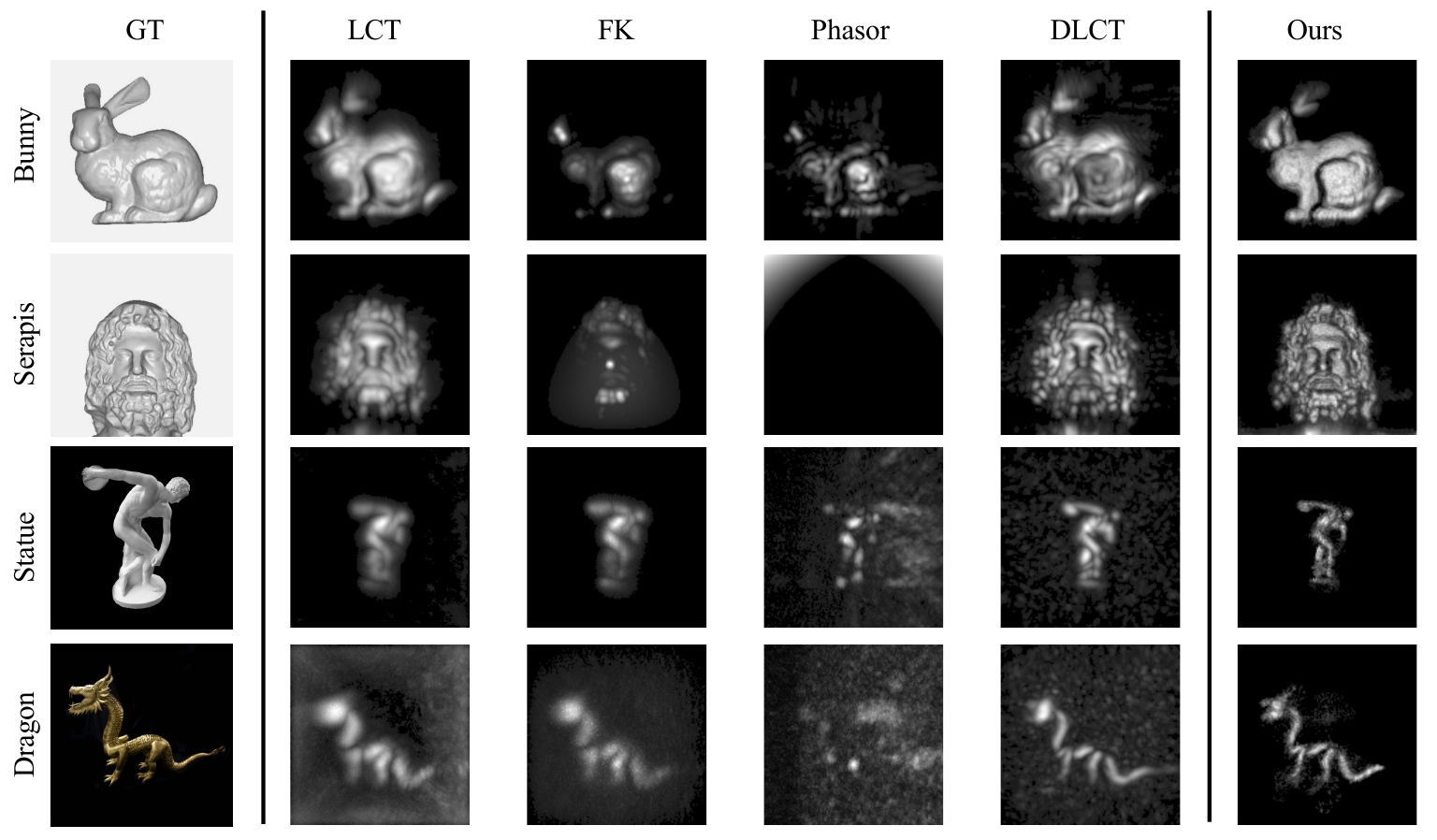}
\caption{Additional comparisons with FFT-based methods, which employs the bicubic interpolation as an additional technique to upsample the input transients to $128 \times 128$ scanning resolution.}
\label{fig:supp_comp_bicubic}
\end{figure*}

%% file: Supple/Partials/experiment_various_resolution.tex
\begin{figure*}[t]
\centering

\includegraphics[trim={0 0 0 0}, width=1\columnwidth]{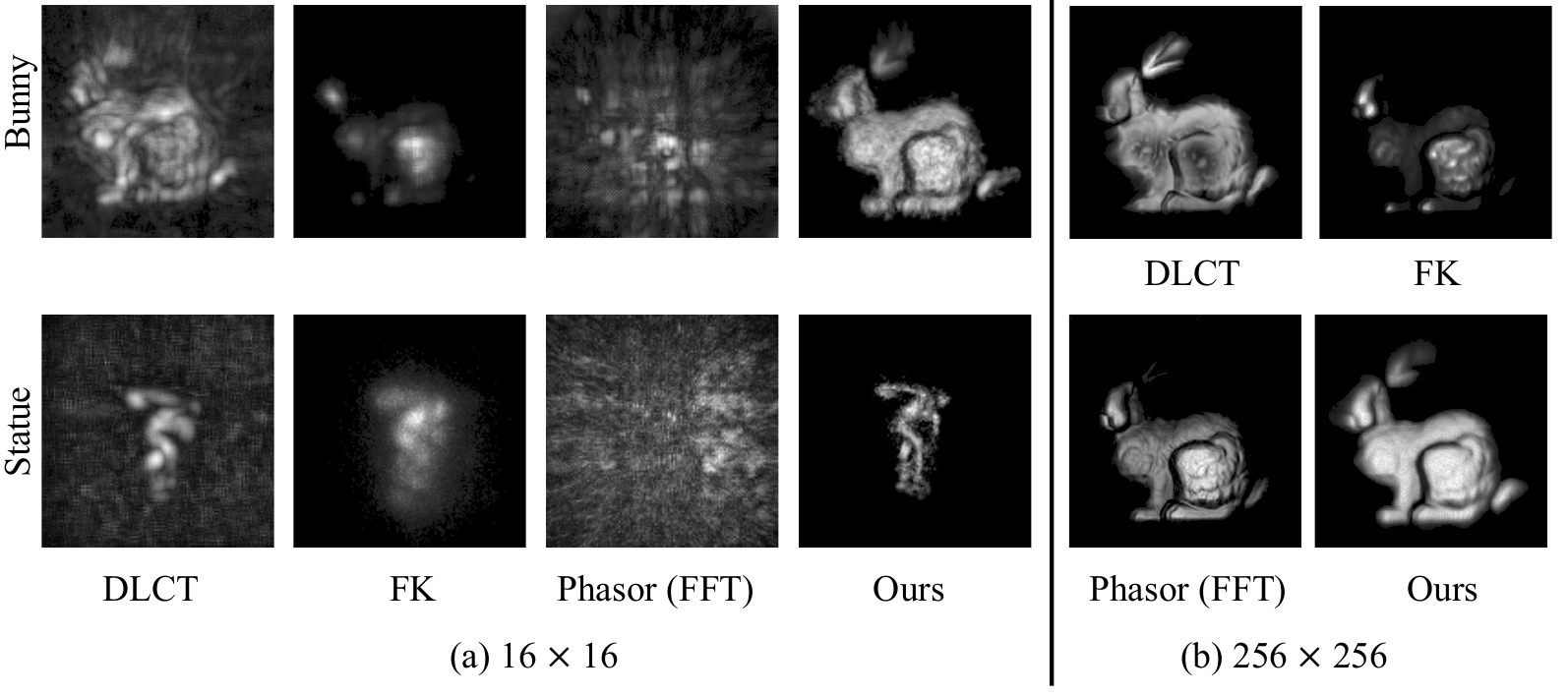}

\caption{\textbf{(a)} Results on $16\times16$ samplings, upsampled to $128 \times 128$ by filling unsampled pixels with zero. FFT-based methods produce unfavorable results with artifacts regardless of the upsampling techniques. \textbf{(b)} Results on $256 \times 256$ transients of ZNLOS Bunny.
}
\label{fig:various_resolution_in_rebuttal}
\end{figure*}

%% file: Supple/Partials/table_runtime_all.tex
\begin{table}[t]
\setlength{\tabcolsep}{5pt}
\centering
\small

\caption{The ratio of active regions and the reconstruction time of all instances. The reconstruction time is measured using a single commercial RTX 3090 GPU. While the reconstruction time varies across the instances, our method typically takes about a minute to reconstruct $128 \times 128$ hidden volumes.}

\begin{tabular}{c|ccc}
\toprule
Scene & num. iter & active ratio & recon. time \\
\midrule
Statue & 1,000 & 0.9 \% & 25 s \\
Dragon & 1,000 & 2.6 \% & 65 s \\
Bunny & 1,000 & 3.0 \% & 54 s \\
Serapis & 1,000 & 8.2 \% & 130 s \\
Bunny (non-confocal) & 1,000 & 3.3 \% & 91 s \\
NT (non-planar) & 1,000 & 0.4 \% & 70 s \\
\bottomrule
\end{tabular}

\label{table:supp_runtime_all}
\vspace{-5pt}
\end{table}

%% file: Supple/Partials/experiment_fbp_pf.tex
\begin{figure*}[t]
\centering
\includegraphics[trim={0 0 0 0}, width=1\linewidth]{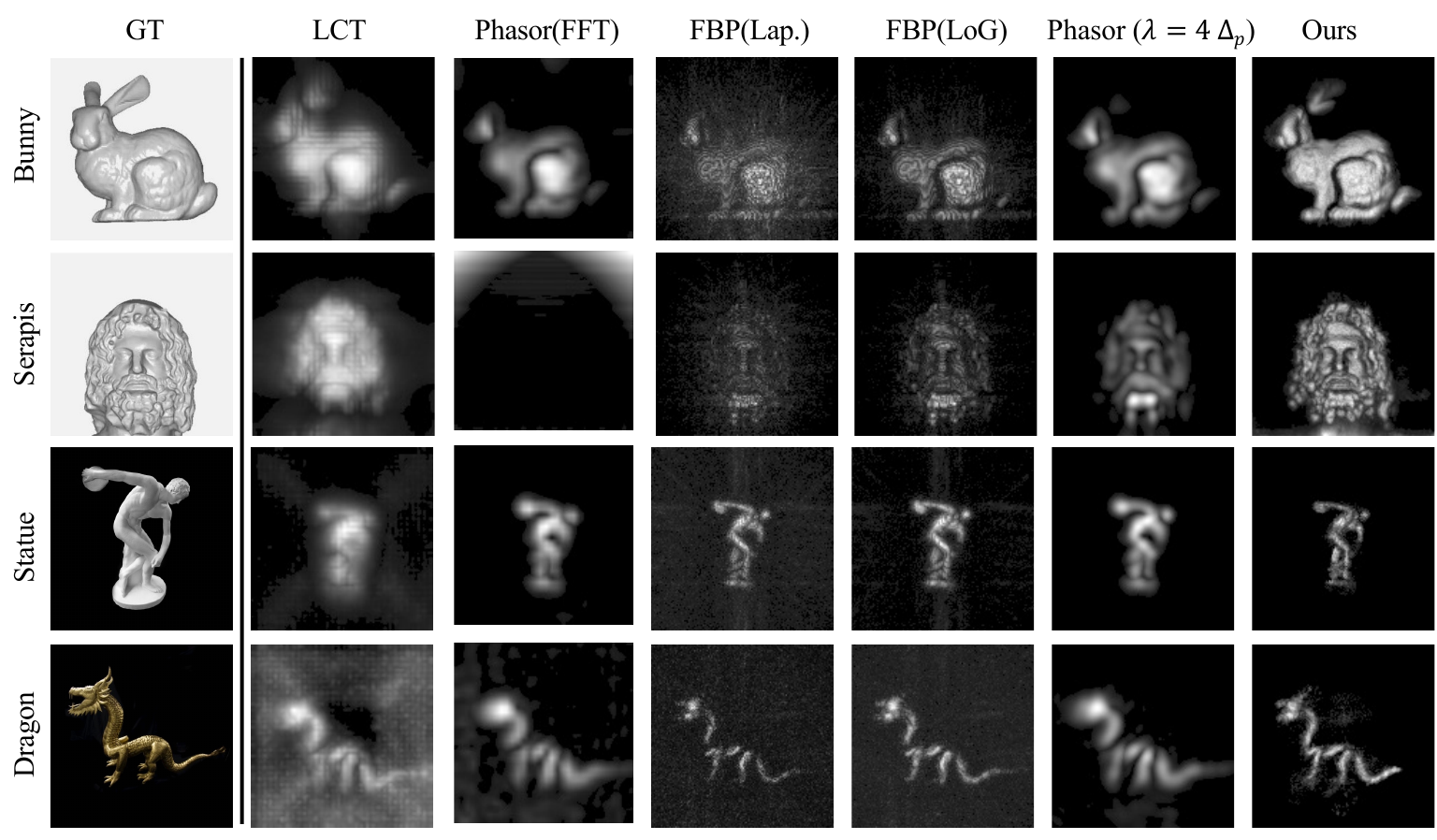}
\caption{Additional comparisons including LCT, Phasor(FFT-based), FBP with a Laplacian filter (Lap.), FBP with a Laplacian-of-Gaussian filter (LoG), and Phasor field with wavelengths $\lambda = 2\Delta_p$ where $\Delta_p$ is the sampling distance.}
\label{fig:supp_comp_bp}
\end{figure*}

%% file: Supple/Partials/experiment_normal_map.tex
\begin{figure*}[t]
\centering

\includegraphics[trim={0 0 0 0}, width=1\columnwidth]{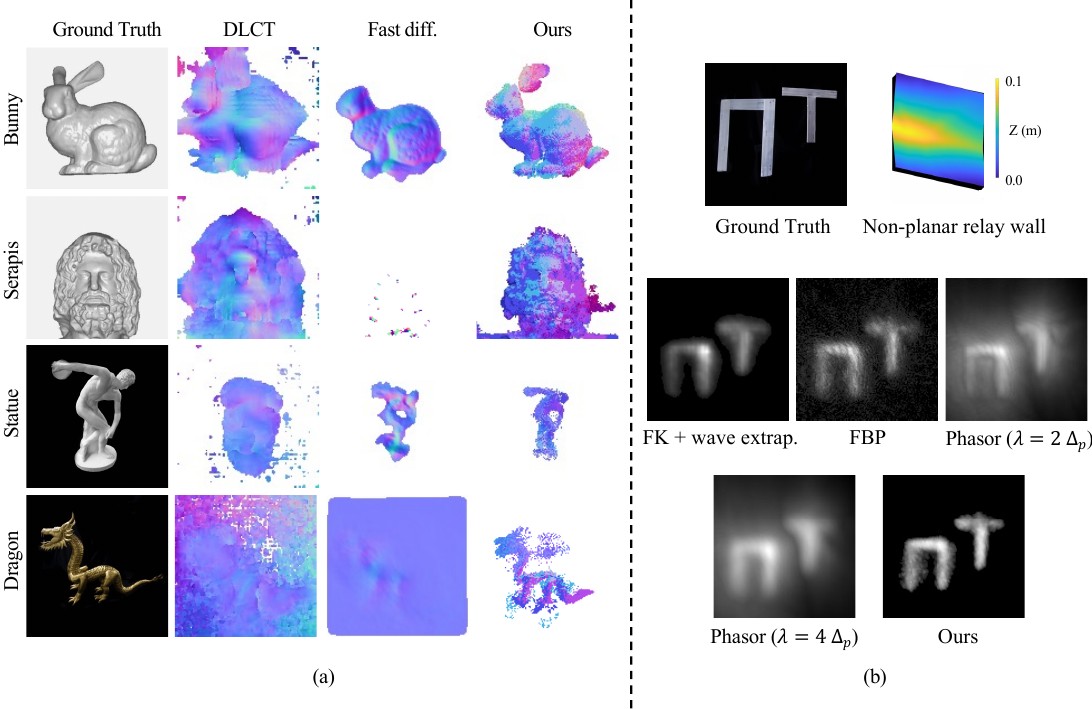}

\caption{\textbf{(a)} Reconstruction results of normal maps. We compare the results with DLCT \cite{young2020dlct} and the fast differentiable renderer proposed by Plack \etal \cite{plack2023fast} (denoted as Fast diff.). \textbf{(b)} Reconstruction results with non-planar relay walls. We additionally compare the results with Phasor field with a BP solver, using the wavelengths $\lambda = 2\Delta_p$ and $\lambda = 4\Delta_p$, where $\Delta_p$ is the sampling distance. }
\label{fig:normal_map_and_non_planar}
\end{figure*}

%% file: Supple/Partials/experiment_each_epoch.tex
\begin{figure*}[t]
\centering
\includegraphics[trim={0 0 0 0}, width=1\linewidth]{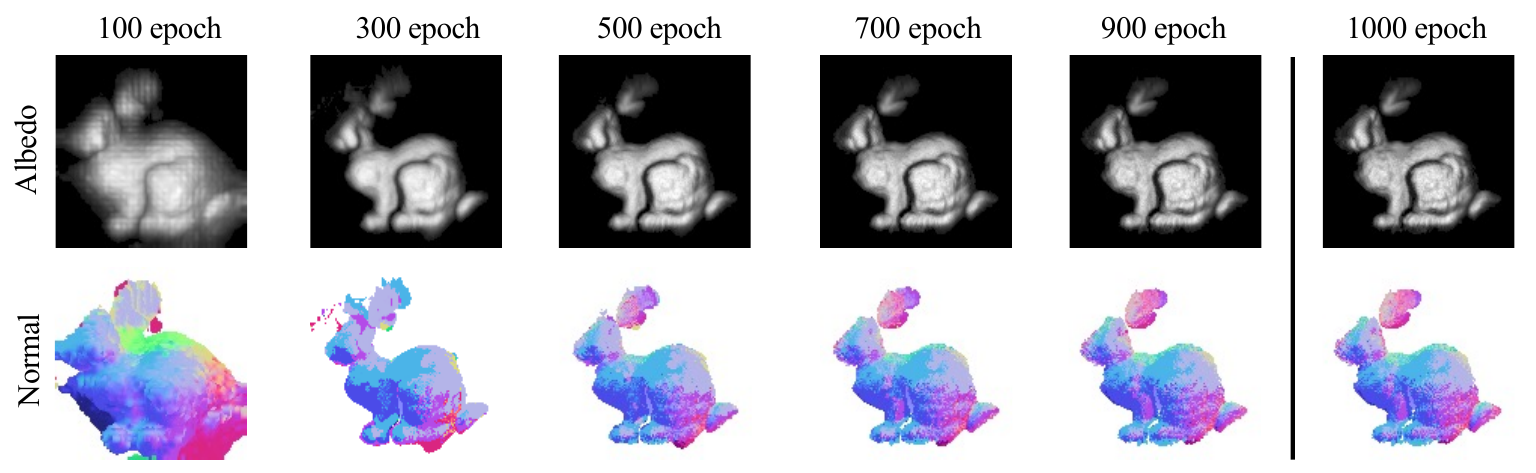}
\caption{Results at several step during the optimization.}
\label{fig:supp_each_epoch}
\end{figure*}

%% file: Supple/Partials/experiment_depth_map_error.tex
\begin{figure*}[t]
\centering
\includegraphics[trim={0 0 0 0}, width=1\linewidth]{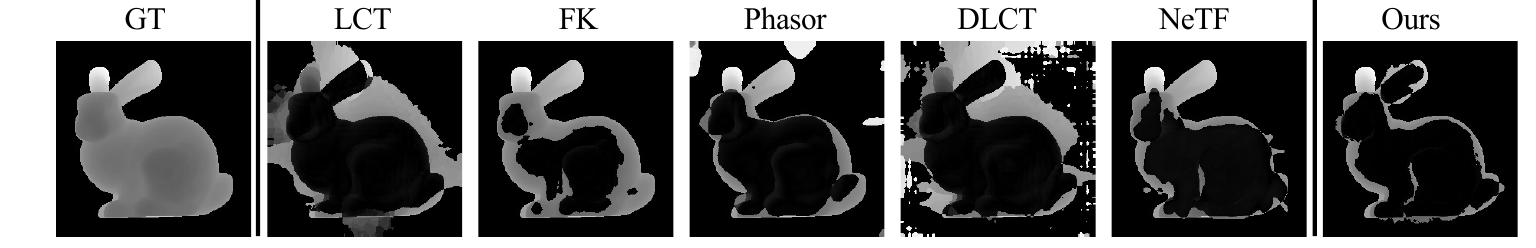}
\caption{Error map visualizations of the reconstructed depth maps on ZNLOS \cite{galindo19znlos} Bunny.}
\label{fig:supp_depth_map}
\end{figure*}

%% file: Supple/Partials/experiment_normal_reconstruction.tex
\begin{figure*}[t]
\centering
\includegraphics[trim={0 0 0 0}, width=1\linewidth]{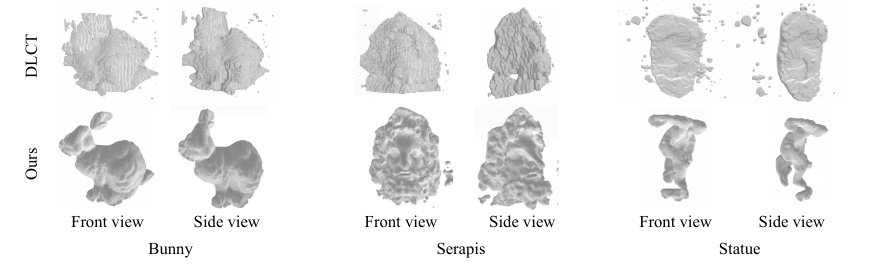}

\caption{Surface reconstruction results. We compare the reconstructed surfaces with DLCT \cite{young2020dlct}.}
\label{fig:surface_reconstruction}
\end{figure*}

%% file: Supple/Partials/albation_continuous_sampling.tex
\begin{figure}[t]
\centering
\includegraphics[trim={0 0 0 0}, width=1\linewidth]{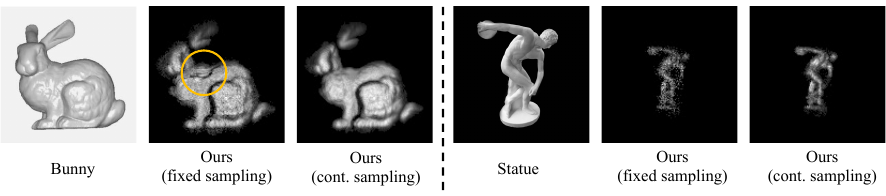}
\caption{Ablation results on the continuous point sampling. We deliver the results with the continuous (denoted as cont.) and the fixed point (denoted as fixed) sampling.}

\label{fig:ablation_continous_sampling}
\end{figure}

%% file: Supple/Partials/ablation_exposure.tex
\begin{figure*}[t]
\centering
\includegraphics[trim={0 0 0 0}, width=1\linewidth]{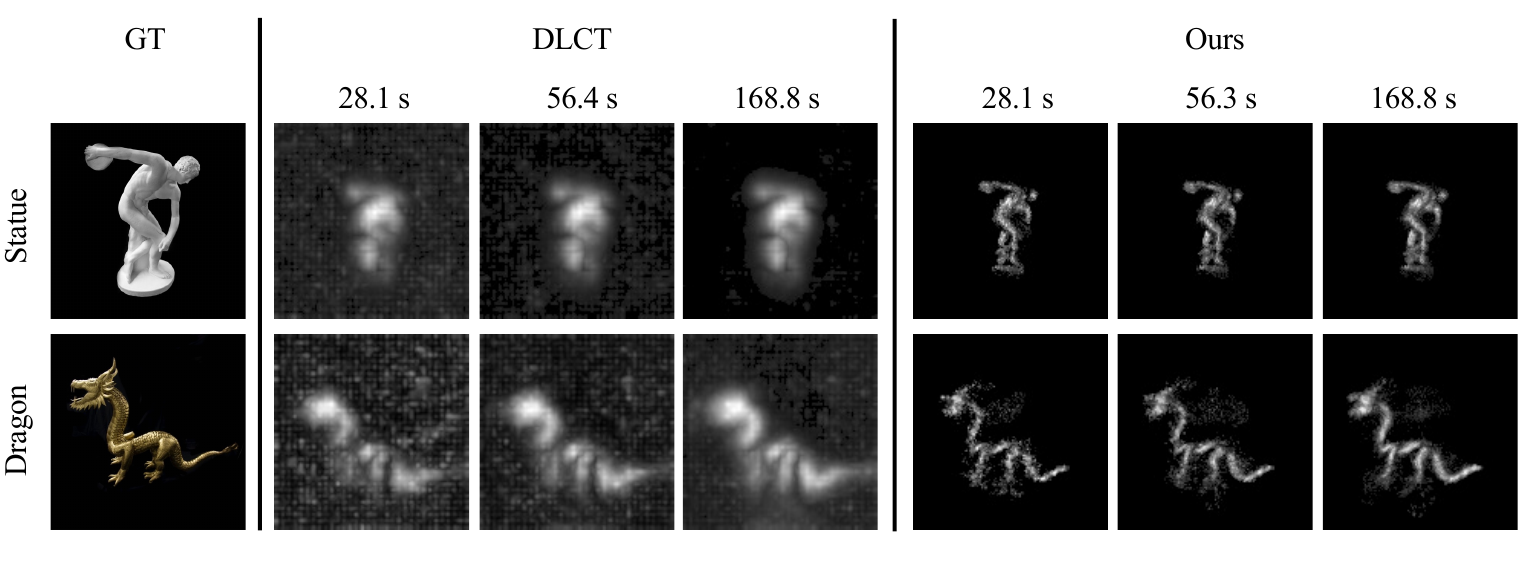}
\caption{Results on the confocal real-world measurements \cite{lindell2019fk} with various exposure time. We use the measurements with $28.1\ \textrm{s}$, $56.4\ \textrm{s}$, $168.8\ \textrm{s}$ total exposure time, which correspond to 30, 60, 180 minute total exposure time of the original measurements.}

\label{fig:ablation_exposure}
\end{figure*}

%% file: main_arxiv.bbl
\begin{thebibliography}{10}
\providecommand{\url}[1]{\texttt{#1}}
\providecommand{\urlprefix}{URL }
\providecommand{\doi}[1]{https://doi.org/#1}

\bibitem{ahn2019convolutional}
Ahn, B., Dave, A., Veeraraghavan, A., Gkioulekas, I., Sankaranarayanan, A.C.: Convolutional approximations to the general non-line-of-sight imaging operator. In: Proceedings of the IEEE/CVF International Conference on Computer Vision (ICCV). pp. 7889--7899 (2019)

\bibitem{aittala2019mirrors}
Aittala, M., Sharma, P., Murmann, L., Yedidia, A., Wornell, G., Freeman, B., Durand, F.: Computational mirrors: Blind inverse light transport by deep matrix factorization. Advances in Neural Information Processing Systems (NeurIPS)  \textbf{32} (2019)

\bibitem{arellano2017fast}
Arellano, V., Gutierrez, D., Jarabo, A.: Fast back-projection for non-line of sight reconstruction. Optics express  \textbf{25}(10),  11574--11583 (2017)

\bibitem{batarseh2018passive}
Batarseh, M., Sukhov, S., Shen, Z., Gemar, H., Rezvani, R., Dogariu, A.: Passive sensing around the corner using spatial coherence. Nature communications  \textbf{9}(1), ~1--6 (2018)

\bibitem{boger2019passive}
Boger-Lombard, J., Katz, O.: Passive optical time-of-flight for non line-of-sight localization. Nature communications  \textbf{10}(1), ~1--9 (2019)

\bibitem{bouman2017turning}
Bouman, K.L., Ye, V., Yedidia, A.B., Durand, F., Wornell, G.W., Torralba, A., Freeman, W.T.: Turning corners into cameras: Principles and methods. In: Proceedings of the IEEE International Conference on Computer Vision. pp. 2270--2278 (2017)

\bibitem{chen2020lfe}
Chen, W., Wei, F., Kutulakos, K.N., Rusinkiewicz, S., Heide, F.: Learned feature embeddings for non-line-of-sight imaging and recognition. ACM Transactions on Graphics (TOG)  \textbf{39}(6),  1--18 (2020)

\bibitem{choi2023self}
Choi, K., Kim, I., Choi, D., Marco, J., Gutierrez, D., Kim, M.H.: Self-calibrating, fully differentiable nlos inverse rendering. In: SIGGRAPH Asia 2023 Conference Papers. pp. 1--11 (2023)

\bibitem{chopite2020deep}
Chopite, J.G., Hullin, M.B., Wand, M., Iseringhausen, J.: Deep non-line-of-sight reconstruction. In: Proceedings of the IEEE/CVF Conference on Computer Vision and Pattern Recognition (CVPR). pp. 960--969 (2020)

\bibitem{clevert2015fast}
Clevert, D.A., Unterthiner, T., Hochreiter, S.: Fast and accurate deep network learning by exponential linear units (elus). arXiv preprint arXiv:1511.07289  (2015)

\bibitem{galindo19znlos}
Galindo, M., Marco, J., O'Toole, M., Wetzstein, G., Gutierrez, D., Jarabo, A.: A dataset for benchmarking time-resolved non-line-of-sight imaging (2019), \url{https://graphics.unizar.es/nlos}

\bibitem{grau2022occlusion}
Grau, J., Plack, M., Haehn, P., Weinmann, M., Hullin, M.: Occlusion fields: An implicit representation for non-line-of-sight surface reconstruction. arXiv preprint arXiv:2203.08657  (2022)

\bibitem{gu2023fast}
Gu, C., Sultan, T., Masumnia-Bisheh, K., Waller, L., Velten, A.: Fast non-line-of-sight imaging with non-planar relay surfaces. In: 2023 IEEE International Conference on Computational Photography (ICCP). pp. 1--12. IEEE (2023)

\bibitem{heide2019partial}
Heide, F., O’Toole, M., Zang, K., Lindell, D.B., Diamond, S., Wetzstein, G.: Non-line-of-sight imaging with partial occluders and surface normals. ACM Transactions on Graphics (TOG)  \textbf{38}(3),  1--10 (2019)

\bibitem{iseringhausen2020non}
Iseringhausen, J., Hullin, M.B.: Non-line-of-sight reconstruction using efficient transient rendering. ACM Transactions on Graphics (ToG)  \textbf{39}(1),  1--14 (2020)

\bibitem{isogawa2020efficient}
Isogawa, M., Chan, D., Yuan, Y., Kitani, K., O’Toole, M.: Efficient non-line-of-sight imaging from transient sinograms. In: Proceedings of the European Conference on Computer Vision (ECCV). pp. 193--208. Springer (2020)

\bibitem{isogawa2020c2nlos}
Isogawa, M., Chan, D., Yuan, Y., Kitani, K., O’Toole, M.: Efficient non-line-of-sight imaging from transient sinograms. In: Proceedings of the European Conference on Computer Vision (ECCV). pp. 193--208. Springer (2020)

\bibitem{isogawa2020optical}
Isogawa, M., Yuan, Y., O'Toole, M., Kitani, K.M.: Optical non-line-of-sight physics-based 3d human pose estimation. In: Proceedings of the IEEE/CVF Conference on Computer Vision and Pattern Recognition (CVPR). pp. 7013--7022 (2020)

\bibitem{kazhdan2006poisson}
Kazhdan, M., Bolitho, M., Hoppe, H.: Poisson surface reconstruction. In: Proceedings of the fourth Eurographics symposium on Geometry processing. vol.~7 (2006)

\bibitem{kingma2014adam}
Kingma, D.P., Ba, J.: Adam: A method for stochastic optimization. arXiv preprint arXiv:1412.6980  (2014)

\bibitem{kirmani2009looking}
Kirmani, A., Hutchison, T., Davis, J., Raskar, R.: Looking around the corner using transient imaging. In: Proceedings of the IEEE/CVF International Conference on Computer Vision (ICCV). pp. 159--166 (2009)

\bibitem{la2020non}
La~Manna, M., Nam, J.H., Reza, S.A., Velten, A.: Non-line-of-sight-imaging using dynamic relay surfaces. Optics express  \textbf{28}(4),  5331--5339 (2020)

\bibitem{lindell2019acoustic}
Lindell, D.B., Wetzstein, G., Koltun, V.: Acoustic non-line-of-sight imaging. In: Proceedings of the IEEE/CVF Conference on Computer Vision and Pattern Recognition (CVPR). pp. 6780--6789 (2019)

\bibitem{lindell2019fk}
Lindell, D.B., Wetzstein, G., O'Toole, M.: Wave-based non-line-of-sight imaging using fast fk migration. ACM Transactions on Graphics (TOG)  \textbf{38}(4),  1--13 (2019)

\bibitem{liu2020nsvf}
Liu, L., Gu, J., Lin, K.Z., Chua, T.S., Theobalt, C.: Neural sparse voxel fields. NeurIPS  (2020)

\bibitem{liu2019analysis}
Liu, X., Bauer, S., Velten, A.: Analysis of feature visibility in non-line-of-sight measurements. In: Proceedings of the IEEE/CVF Conference on Computer Vision and Pattern Recognition (CVPR). pp. 10140--10148 (2019)

\bibitem{liu2020diffraction}
Liu, X., Bauer, S., Velten, A.: Phasor field diffraction based reconstruction for fast non-line-of-sight imaging systems. Nature communications  \textbf{11}(1),  1--13 (2020)

\bibitem{liu2019phasor}
Liu, X., Guill{\'e}n, I., La~Manna, M., Nam, J.H., Reza, S.A., Huu~Le, T., Jarabo, A., Gutierrez, D., Velten, A.: Non-line-of-sight imaging using phasor-field virtual wave optics. Nature  \textbf{572}(7771),  620--623 (2019)

\bibitem{liu2023sscr}
Liu, X., Wang, J., Xiao, L., Fu, X., Qiu, L., Shi, Z.: Few-shot non-line-of-sight imaging with signal-surface collaborative regularization. In: Proceedings of the IEEE/CVF Conference on Computer Vision and Pattern Recognition (CVPR). pp. 13303--13312 (2023)

\bibitem{martin2021nerf}
Martin-Brualla, R., Radwan, N., Sajjadi, M.S., Barron, J.T., Dosovitskiy, A., Duckworth, D.: Nerf in the wild: Neural radiance fields for unconstrained photo collections. In: Proceedings of the IEEE/CVF Conference on Computer Vision and Pattern Recognition (CVPR). pp. 7210--7219 (2021)

\bibitem{mildenhall2020nerf}
Mildenhall, B., Srinivasan, P.P., Tancik, M., Barron, J.T., Ramamoorthi, R., Ng, R.: Nerf: Representing scenes as neural radiance fields for view synthesis. In: Proceedings of the European Conference on Computer Vision (ECCV). pp. 405--421. Springer (2020)

\bibitem{mu2022rescue}
Mu, F., Mo, S., Peng, J., Liu, X., Nam, J.H., Raghavan, S., Velten, A., Li, Y.: Physics to the rescue: Deep non-line-of-sight reconstruction for high-speed imaging. IEEE Transactions on Pattern Analysis and Machine Intelligence (PAMI)  (2022)

\bibitem{nam2021low}
Nam, J.H., Brandt, E., Bauer, S., Liu, X., Renna, M., Tosi, A., Sifakis, E., Velten, A.: Low-latency time-of-flight non-line-of-sight imaging at 5 frames per second. Nature communications  \textbf{12}(1), ~6526 (2021)

\bibitem{o2018lct}
O’Toole, M., Lindell, D.B., Wetzstein, G.: Confocal non-line-of-sight imaging based on the light-cone transform. Nature  \textbf{555}(7696),  338--341 (2018)

\bibitem{pei2021dynamic}
Pei, C., Zhang, A., Deng, Y., Xu, F., Wu, J., David, U., Li, L., Qiao, H., Fang, L., Dai, Q.: Dynamic non-line-of-sight imaging system based on the optimization of point spread functions. Optics Express  \textbf{29}(20),  32349--32364 (2021)

\bibitem{plack2023fast}
Plack, M., Callenberg, C., Schneider, M., Hullin, M.B.: Fast differentiable transient rendering for non-line-of-sight reconstruction. In: Proceedings of the IEEE/CVF Winter Conference on Applications of Computer Vision. pp. 3067--3076 (2023)

\bibitem{rapp2020edge}
Rapp, J., Saunders, C., Tachella, J., Murray-Bruce, J., Altmann, Y., Tourneret, J.Y., McLaughlin, S., Dawson, R.M., Wong, F.N., Goyal, V.K.: Seeing around corners with edge-resolved transient imaging. Nature communications  \textbf{11}(1), ~5929 (2020)

\bibitem{saunders2019computational}
Saunders, C., Murray-Bruce, J., Goyal, V.K.: Computational periscopy with an ordinary digital camera. Nature  \textbf{565}(7740),  472--475 (2019)

\bibitem{seidel2020two}
Seidel, S.W., Murray-Bruce, J., Ma, Y., Yu, C., Freeman, W.T., Goyal, V.K.: Two-dimensional non-line-of-sight scene estimation from a single edge occluder. IEEE Transactions on Computational Imaging  \textbf{7},  58--72 (2020)

\bibitem{sharma2021what}
Sharma, P., Aittala, M., Schechner, Y.Y., Torralba, A., Wornell, G.W., Freeman, W.T., Durand, F.: What you can learn by staring at a blank wall. In: Proceedings of the IEEE/CVF International Conference on Computer Vision (ICCV). pp. 2330--2339 (2021)

\bibitem{shen2021netf}
Shen, S., Wang, Z., Liu, P., Pan, Z., Li, R., Gao, T., Li, S., Yu, J.: Non-line-of-sight imaging via neural transient fields. IEEE Transactions on Pattern Analysis and Machine Intelligence (PAMI)  \textbf{43}(7),  2257--2268 (2021)

\bibitem{shi2024passive}
Shi, X., Tang, M., Zhang, S., Qiao, K., Gao, X., Jin, C.: Passive localization and reconstruction of multiple non-line-of-sight objects in a scene with a large visible transmissive window. Optics Express  \textbf{32}(6),  10104--10118 (2024)

\bibitem{tanaka2020polarized}
Tanaka, K., Mukaigawa, Y., Kadambi, A.: Polarized non-line-of-sight imaging. In: Proceedings of the IEEE/CVF Conference on Computer Vision and Pattern Recognition (CVPR). pp. 2136--2145 (2020)

\bibitem{tsai2017first}
Tsai, C.Y., Kutulakos, K.N., Narasimhan, S.G., Sankaranarayanan, A.C.: The geometry of first-returning photons for non-line-of-sight imaging. In: Proceedings of the IEEE/CVF Conference on Computer Vision and Pattern Recognition (CVPR). pp. 7216--7224 (2017)

\bibitem{tsai2019optimization}
Tsai, C.Y., Sankaranarayanan, A.C., Gkioulekas, I.: Beyond volumetric albedo--a surface optimization framework for non-line-of-sight imaging. In: Proceedings of the IEEE/CVF Conference on Computer Vision and Pattern Recognition (CVPR). pp. 1545--1555 (2019)

\bibitem{velten2012fbp}
Velten, A., Willwacher, T., Gupta, O., Veeraraghavan, A., Bawendi, M.G., Raskar, R.: Recovering three-dimensional shape around a corner using ultrafast time-of-flight imaging. Nature communications  \textbf{3}(1), ~1--8 (2012)

\bibitem{wang2023non}
Wang, J., Liu, X., Xiao, L., Shi, Z., Qiu, L., Fu, X.: Non-line-of-sight imaging with signal superresolution network. In: Proceedings of the IEEE/CVF Conference on Computer Vision and Pattern Recognition. pp. 17420--17429 (2023)

\bibitem{wu2021over}
Wu, C., Liu, J., Huang, X., Li, Z.P., Yu, C., Ye, J.T., Zhang, J., Zhang, Q., Dou, X., Goyal, V.K., et~al.: Non--line-of-sight imaging over 1.43 km. Proceedings of the National Academy of Sciences  \textbf{118}(10),  e2024468118 (2021)

\bibitem{xin2019fermat}
Xin, S., Nousias, S., Kutulakos, K.N., Sankaranarayanan, A.C., Narasimhan, S.G., Gkioulekas, I.: A theory of fermat paths for non-line-of-sight shape reconstruction. In: Proceedings of the IEEE/CVF Conference on Computer Vision and Pattern Recognition (CVPR). pp. 6800--6809 (2019)

\bibitem{ye2021compressed}
Ye, J.T., Huang, X., Li, Z.P., Xu, F.: Compressed sensing for active non-line-of-sight imaging. Optics Express  \textbf{29}(2),  1749--1763 (2021)

\bibitem{yedidia2019using}
Yedidia, A.B., Baradad, M., Thrampoulidis, C., Freeman, W.T., Wornell, G.W.: Using unknown occluders to recover hidden scenes. In: Proceedings of the IEEE/CVF Conference on Computer Vision and Pattern Recognition (CVPR). pp. 12231--12239 (2019)

\bibitem{young2020dlct}
Young, S.I., Lindell, D.B., Girod, B., Taubman, D., Wetzstein, G.: Non-line-of-sight surface reconstruction using the directional light-cone transform. In: Proceedings of the IEEE/CVF Conference on Computer Vision and Pattern Recognition (CVPR). pp. 1407--1416 (2020)

\bibitem{yu2021plenoctrees}
Yu, A., Li, R., Tancik, M., Li, H., Ng, R., Kanazawa, A.: Plenoctrees for real-time rendering of neural radiance fields. In: Proceedings of the IEEE/CVF International Conference on Computer Vision. pp. 5752--5761 (2021)

\bibitem{zhu2022remapping}
Zhu, D., Cai, W.: Fast non-line-of-sight imaging with two-step deep remapping. ACS Photonics  \textbf{9}(6),  2046--2055 (2022)

\end{thebibliography}
